\documentclass{article}
\usepackage{scml2025}
\usepackage{multicol}
\usepackage[utf8]{inputenc} 
\usepackage[T1]{fontenc}    
\usepackage{hyperref}       
\usepackage{url}            
\usepackage{booktabs}       
\usepackage{amsfonts}       
\usepackage{nicefrac}       
\usepackage{microtype}      
\usepackage{lipsum}         
\usepackage{graphicx}
\usepackage[square,sort,comma,numbers]{natbib}
\usepackage{doi}
\usepackage{wrapfig}
\usepackage{float}
\usepackage{enumitem}
\usepackage{subcaption}
\usepackage{amsmath}
\usepackage{cleveref}  
\title{Integrating Large Language Models For Monte Carlo Simulation of Chemical Reaction Networks}
\renewcommand{\shorttitle}{Automating CRN Simulation With LLM}

\date{}

\newif\ifuniqueAffiliation
\uniqueAffiliationtrue

\ifuniqueAffiliation 
\author{%
	Sadikshya Gyawali \thanks{Primary Author} \\
	E.K. Solutions Pvt. Ltd.\\
	Lalitpur, Nepal \\
	\texttt{sadikshya.gyawali@ekbana.info} \\
	\And
        Ashwini Mandal \\
	E.K. Solutions Pvt. Ltd.\\
	Lalitpur, Nepal \\
	\texttt{ashwini.mandal@ekbana.info} \\
        \And
        Manish Dahal \\
	E.K. Solutions Pvt. Ltd.\\
	Lalitpur, Nepal \\
	\texttt{manish.dahal@ekbana.info} \\
        \And
        Manish Awale \\
	E.K. Solutions Pvt. Ltd.\\
	Lalitpur, Nepal \\
	\texttt{manish.awale@ekbana.info} \\
        \And
	Sanjay Rijal \\
        Institut de Física d'Altes Energies (IFAE) \\ 
        Barcilona \\
        \textit{srijal@cern.ch} \\
        \And
	Shital Adhikari \\
        Stevens Institute of Technology \\ 
        New Jersey, USA \\
        \textit{sadhikar1@stevens.edu} \\
    \And
	Vaghawan Ojha \\
        E.K. Solutions Pvt. Ltd. \\ 
        Lalitpur, Nepal\\
	\texttt{vaghawan.ojha@ekbana.net} \\
}
\else
\usepackage{authblk}

\fi

\hypersetup{
pdftitle={Integrating Large Language Models For Monte Carlo Simulation of Chemical Reaction Networks},
pdfsubject={q-bio.NC, q-bio.QM},
pdfauthor={Vaghawan.~Ojha, Sadikshya.~Gyawali, Ashwini.~Mandal},
pdfkeywords={Chemical reaction networks, Stochastic simulation, Large language models},
}

\begin{document}
\maketitle

\begin{abstract}
Chemical reaction network is an important method for modeling and exploring complex biological processes, bio-chemical interactions and the behavior of different dynamics in system biology. But, formulating such reaction kinetics takes considerable time. In this paper, we leverage the efficiency of modern large language models to automate the stochastic monte carlo simulation of chemical reaction networks and enable the simulation through the reaction description provided in the form of natural languages. We also integrate this process into widely used simulation tool Copasi to further give the edge and ease to the modelers and researchers. In this work, we show the efficacy and limitations of the modern large language models to parse and create reaction kinetics for modelling complex chemical reaction processes.
\end{abstract}


\section{Introduction}
Chemical reaction networks \cite{chemical-reaction-network-tutorial, Spotte-Smith2023, exploration-of-chemical-reaction, our-1} are fundamental to understanding complex biochemical systems and material science \cite{Cook2009}. The significance of chemical reaction network is that the complete knowledge of all elementary steps, including intermediates, transition structures, and products, allows for kinetic modeling and the prediction of concentration fluxes through the network \cite{exploration-of-chemical-reaction}. Modeling of such reaction network involves first understanding the inherent reactions, being able to capture all products of the reactions and their probable reaction rates and kinetics. It has been known that stochasticity is the key characteristics of many chemical reactions processes, especially the biochemical ones \cite{biochemical-reaction-networks, our-2} where small perturbation is often random, plays significant roles. Such reactions are analytically intractable \cite{biochemical-reaction-networks}, hence stochastic computational simulation is one of the approach that researchers adopt to simulate the dynamics and explore the aggregate behavior and convergence of the system. But putting such complex biochemical reaction networks is not an easy task, involving the identification of all critical and non-critical species, intermediate steps, reaction kinetics, the rate and parameters of the system, and finally be able to bring forth stoichiometry matrix which can then be used for simulation. This process starts by writing down all possible reactions and their products. 
However, translating natural language descriptions of reaction mechanisms into computational models remains a challenge and takes considerable human effort. This paper presents a framework that combines Large Language Models (LLMs) \cite{naveed2024comprehensiveoverviewlargelanguage} with stochastic simulation algorithms to bridge this gap.
Our system leverages state-of-the-art language models to automatically parse and interpret chemical reaction descriptions, converting them into structured mathematical representations (reaction kinetics and stoichiometry matrix) suitable for simulation. By integrating the Gillespie Stochastic Simulation Algorithm (SSA) \cite{gillespie2007stochastic} with modern LLM capabilities, this method may enable researchers to directly use natural language descriptions of reactions instead of working on the reaction kinetics, which may otherwise have to be structured differently based on the kind of software being used. This has the potential to save researchers' time and effort by automating the simulation. Our system also produces Copasi model \cite{hoops2006copasi} which can be directly loaded in the Copasi simulation software, and it also executes the simulation programatically using python integration of Copasi \cite{basico}.  

The framework incorporates six key useful properties: (1) a LLM-powered reaction parser that interprets complex chemical descriptions, (2) an automated stoichiometry matrix generator for mathematical representation of reaction networks, (3) Adjustable implementation of the Gillespie algorithm for stochastic simulation of reaction dynamics, (4) Adjustable Monte Carlo Simulation based on the defined reactions, (5) Copasi Model and simulation which can be either directly executed or can be loaded in Copasi software, (6) An evaluator agent which leverages LLM's reasoning capability to evaluate and re-adjust the parsed details as expected by the next step. This reduces the unnecessary distortion and unclear convergence in the systems of reactions. This integration provides further ease to the process of simulation for studying chemical reaction networks ranging from simple equilibrium reactions to complex polymerization processes.

\section{Reaction Dynamics and Population Behavior in Chemical Reaction Networks}

The system of reactions describes the dynamic interactions between the species $A$, $A_2$, $A_3$, and $A_4$, governed by the rate constants $k_1, k_2, k_3$, and $k_4$. The dynamics of the species are represented by the following differential equations:

\textbf{For $A(t)$:}
\[
\frac{dA}{dt} = -k_1 A^2 - k_2 A_2 A - k_3 A_3 A
\]
The population of $A$ decreases over time as it is consumed in three reactions: self-reaction ($A + A$) to form $A_2$, interaction with $A_2$ to produce $A_3$, and reaction with $A_3$ leading to the formation of $A_4$.

\textbf{For $A_2(t)$:}
\[
\frac{dA_2}{dt} = k_1 A^2 + k_2 A_2 A - k_4 A_2
\]
$A_2$ is produced via the self-reaction of $A$ and consumed when reacting with $A$ to form $A_3$. It also undergoes degradation or transformation at a rate proportional to $k_4$, balancing its production and consumption.

\textbf{For $A_3(t)$:}
\[
\frac{dA_3}{dt} = k_2 A_2 A - k_3 A_3 A
\]
$A_3$ is formed when $A_2$ reacts with $A$ and is consumed during its reaction with $A$ to produce $A_4$. The population of $A_3$ depends on the balance of these two processes.

\textbf{For $A_4(t)$:}
\[
\frac{dA_4}{dt} = k_3 A_3 A + k_4 A_2
\]
$A_4$ is the terminal product of the reaction network, accumulated through two pathways: the reaction between $A_3$ and $A$, and the degradation or transformation of $A_2$. It does not participate in further reactions, leading to a consistent increase in its population.

\begin{figure}[H]
	\begin{center}
		\includegraphics[width=0.8\textwidth]{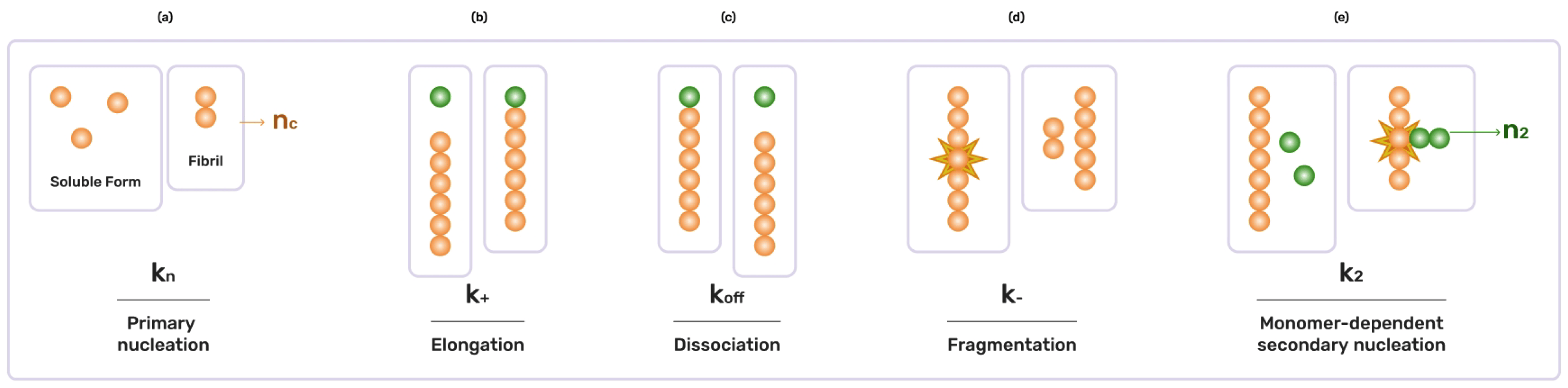}
	\end{center}
	\caption{Example of problem modeled with chemical reaction networks in Amyloid-Beta protein aggregation process (Figure taken from \cite{our-1})}
	\label{fig:example-process}
\end{figure}

This cascade of reactions illustrates a progressive depletion of the primary species $A$, the transient behavior of intermediate species $A_2$ and $A_3$, and the accumulation of the final product of the reaction leading to convergence to some state. 
\subsection{Modeling of Chemical Reaction Networks}
The mathematical modeling of chemical reaction networks involves two primary approaches: deterministic and stochastic. Figure \ref{fig:example-process} shows the chemical reaction network to model the amyloid-beta aggregation process.

\subsubsection*{Deterministic Model}

In the deterministic model, the reactions of $m$ chemical species $X_i$ ($i = 1, 2, \dots, m$) participating in $n$ reactions are characterized as \cite{our-1}:

\[
\sum_{j=1}^m a_{ij} X_j \xrightarrow{k_i} \sum_{j=1}^m b_{ij} X_j, \quad 1 \leq i \leq m,
\]

where $a_{ij}$ and $b_{ij}$ are the stoichiometric coefficients for the reactants and products respectively. These coefficients are non-negative integers. The reaction rates $r_i$, determined by the law of mass action, are given by:

\[
r_i = k_i \prod_{j=1}^m X_j^{a_{ij}},
\]

where $k_i$ is the rate constant of the $i$-th reaction.

By defining the stoichiometric matrix $S$ as:

\[
S = (s_{ij}) = (b_{ij} - a_{ij}),
\]

The system of differential equations describing the concentration dynamics is written as:

\[
\frac{dX}{dt} = Sr,
\]

where $X$ is the vector of species concentrations and $r$ is the vector of reaction rates.

\subsubsection*{Stochastic Model}

In the stochastic model, chemical reaction networks are described using the chemical master equation, which accounts for the probabilistic nature of reactions. The master equation governs the time evolution of the probability distribution $P(X, t)$, where $X(t) = (X_1(t), X_2(t), \dots, X_m(t))$ represents the state of the system at time $t$.

The chemical master equation is expressed as \cite{our-1, biochemical-reaction-networks}:

\[
\frac{dP(X, t)}{dt} = \sum_{i=1}^m P(X - S_i, t) a_i(X - S_i) - \sum_{i=1}^m P(X, t) a_i(X),
\]

where:
\begin{itemize}
    \item $S_i$ is the $i$-th row of the stoichiometric matrix $S$,
    \item $a_i(X)$ is the propensity function of the $i$-th reaction, given by $a_i(X) = k_i \prod_{j=1}^m X_j^{a_{ij}}$.
\end{itemize}

Such stochastic framework ensures that the principle of detailed balance is maintained, equalizing the forward and reverse reaction rates and ensuring the natural phenomenon observed in different chemical reactions.

Such reactions are widely found in chemical reaction networks  \cite{biochemical-reaction-networks, chemical-reaction-network-tutorial, exploration-of-chemical-reaction}, \cite{our-1, nature-oligomers, Ding2024, our-2} uses such kinetics to model the protein aggregation phenomenon in biological conditions. Similar reaction networks have been studied \cite{exploration-of-chemical-reaction} on many natural science applications. Considering the importance of such reactions, and the time it takes to computationally simulate them, we consider to leverage large language models capability to automatically parse the description of the dynamics, create reaction kinetics, and write stoichiometry matrix (which describes the reactions and its product in matrix form). After then, the adjustable Monte Carlo Simulation can finally simulate the dynamics and an Analyzer agent can take the results from the simulation and produce meaningful plots that shows the state and the dynamics of chemical species. Copasi \cite{hoops2006copasi} is a popular simulation software that is used to model chemical reaction dynamics using both deterministic as well as stochastic models. To further ease the process of modeling, our framework also gives user the ability to directly integrate the designed simulation into Copasi, either as a python integration \cite{basico}, or as an importable Copasi model. 

We validated this approach in a number of different reaction kinetics and their dynamics reported in the respective research papers, and results show that our LLM-backed framework can successfully replicate such systems without explicitly having to define the reaction kinetics. Our approach was able to simulate even the complex reactions which involved 56 different reactions including both aggregation and fragmentation, with the stoichiometry matrix of dimension $54 \times 8$.

\section{Proposed Solution}

\begin{figure}[H]
	\begin{center}
		\includegraphics[width=0.8\textwidth]{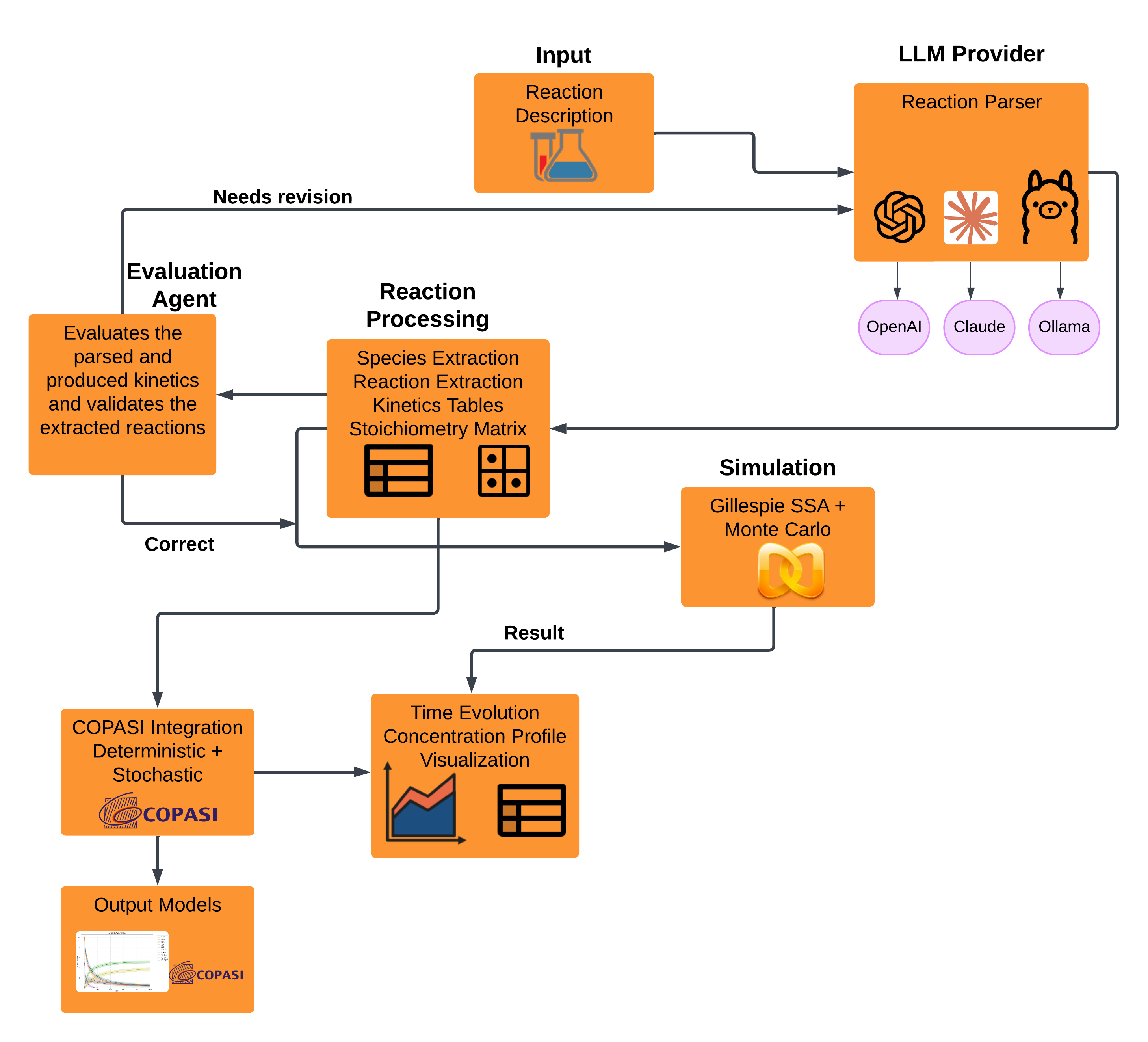}
	\end{center}
	\caption{Architecture of Adopted Method}
	\label{fig:proposed-solution}
\end{figure}

Figure \ref{fig:proposed-solution} depicts the overall approach that was adopted in this study. Instead of kinetic tables or derived stoichiometry matrix, the first step takes the input that contains the description of the reactions in natural language. It further passes to the reaction parser, which extracts the species involved in the reaction and identifies the reaction constants associated with each reaction, creates kinetic tables and turns it into the stoichiometry matrix that can be a direct input to the next step. After the details are successfully extracted, the matrix is passed to Gillespie algorithm which is wrapped by the Monte Carlo (MC) Simulation, different parameters needed for MC are dynamically adjusted and decided by the underlying process. If Copasi model is also choosen, then from the same description, our method will create a Copasi model, run the simulation and also output the concentration of the species after the simulation is finished.  
\section{Results}
\begin{figure}[H]
    \centering

    \begin{subfigure}{0.47\textwidth}
        \centering
        \includegraphics[width=\textwidth]{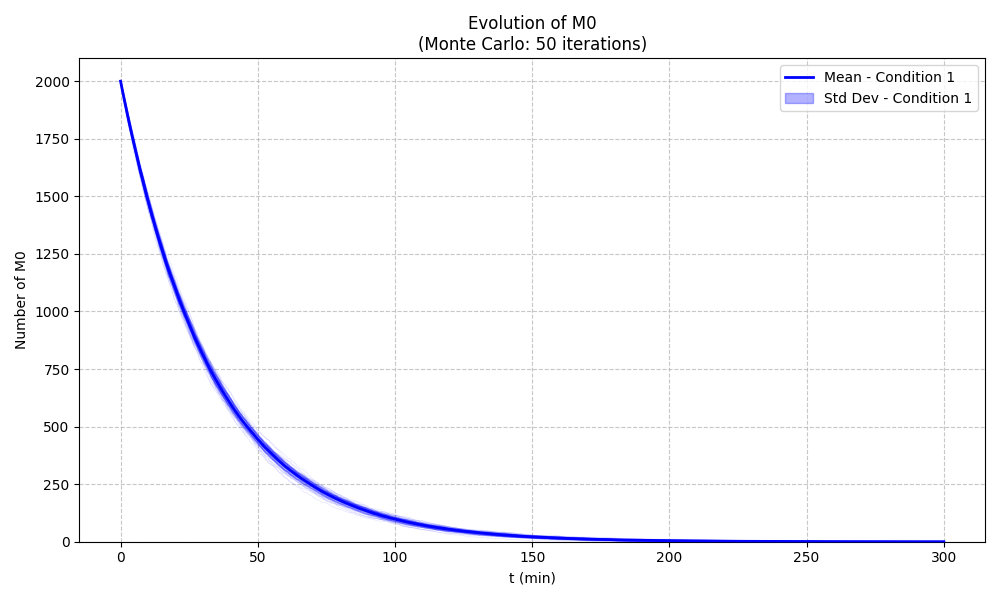}
        \caption{Result of $M_0$}
        \label{fig:m0-nature2}
    \end{subfigure}
    \hfill\hfill\hfill
    \begin{subfigure}{0.47\textwidth}
        \centering
        \includegraphics[width=\textwidth]{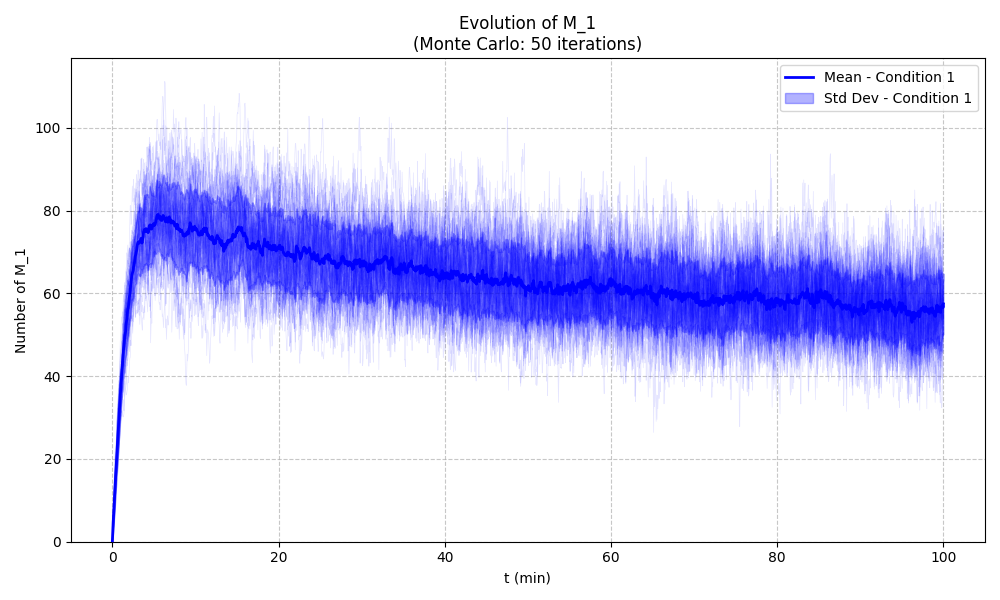}
        \caption{Result of $M_1$}
        \label{fig:m1-nature-2}
    \end{subfigure}

    \vspace{0.5cm} 

    \begin{subfigure}{1.0\textwidth}
        \centering
        \begin{subfigure}{0.475\textwidth}
            \includegraphics[width=\textwidth, height = 4.5cm]{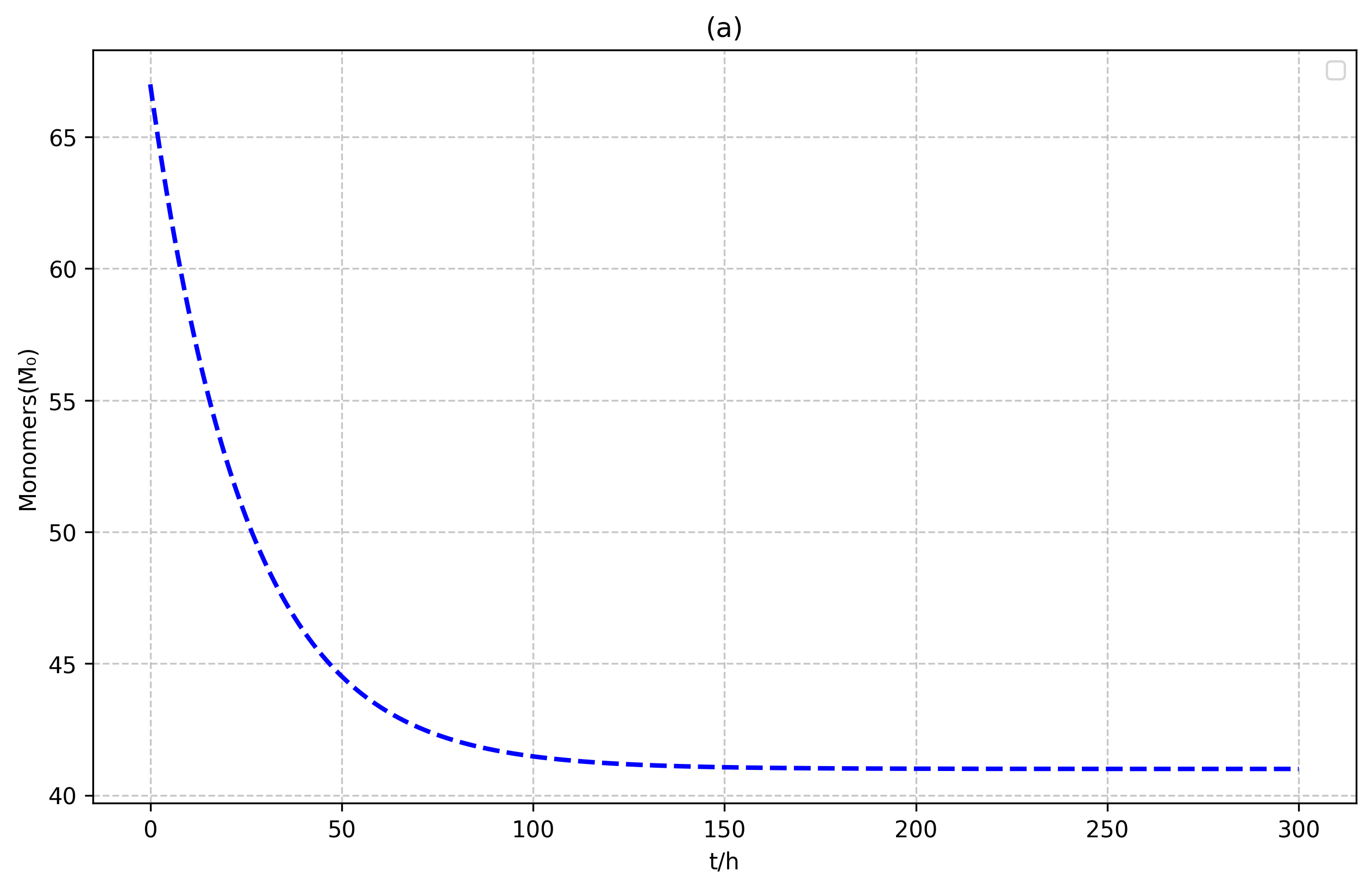}
            \caption{Original result of $M_0$}
        \end{subfigure}
        \hfill\hfill\hfill
        \begin{subfigure}{0.475\textwidth}
            \includegraphics[width=\textwidth, height = 4.5cm]{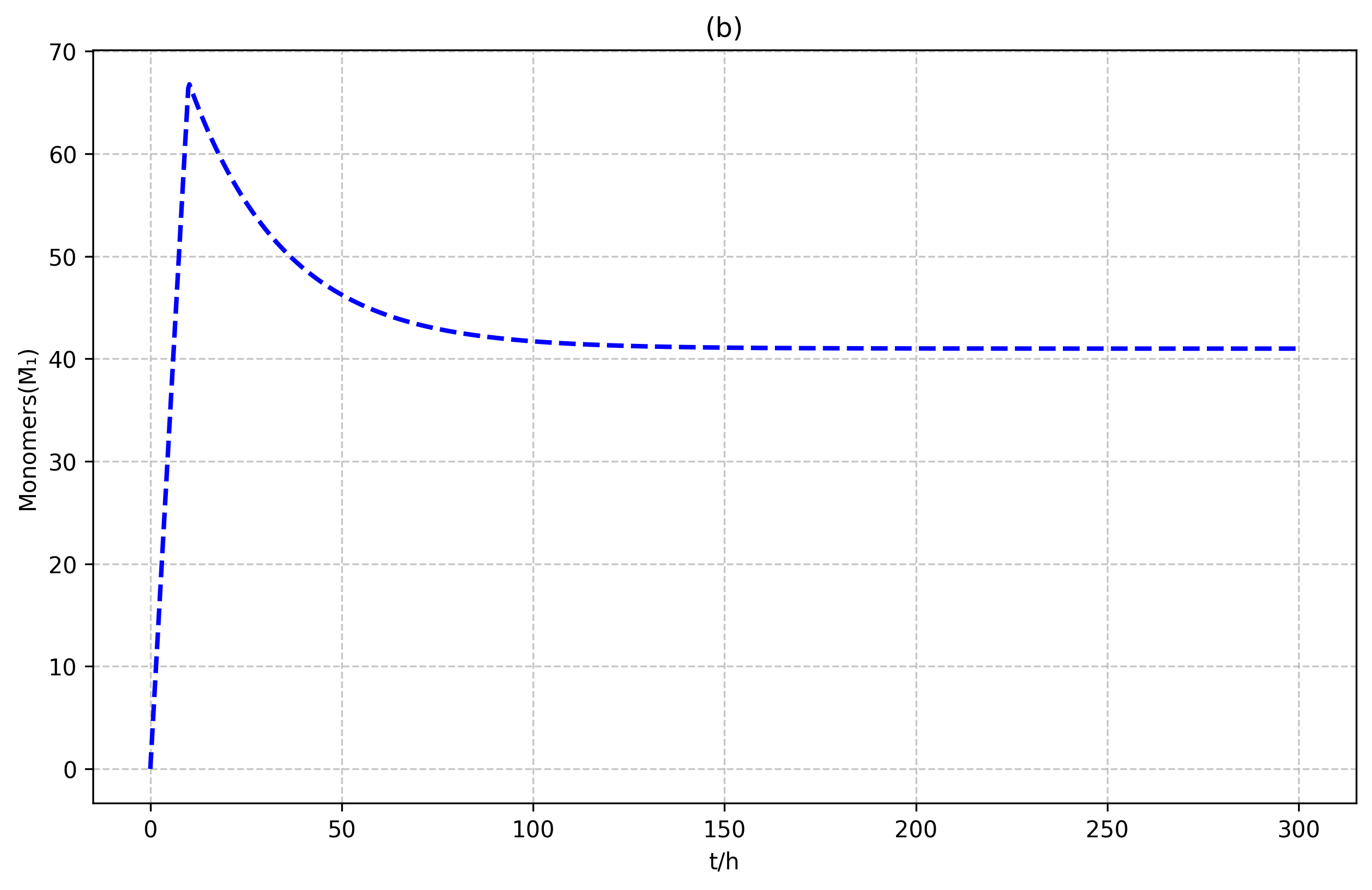}
            \caption{Original result of $M_1$}
        \end{subfigure}
        \label{fig:m1-nature-2-original}
    \end{subfigure}

    \vspace{0.5cm}
    \caption{Comparison of the results from our method with the system described in \cite{Ding2024}, as shown in Figure 5}
    \label{fig:paper-45-results}
\end{figure}

\begin{figure}[H]
    \centering

    \begin{subfigure}{0.47\textwidth}
        \centering
        \includegraphics[width=\textwidth]{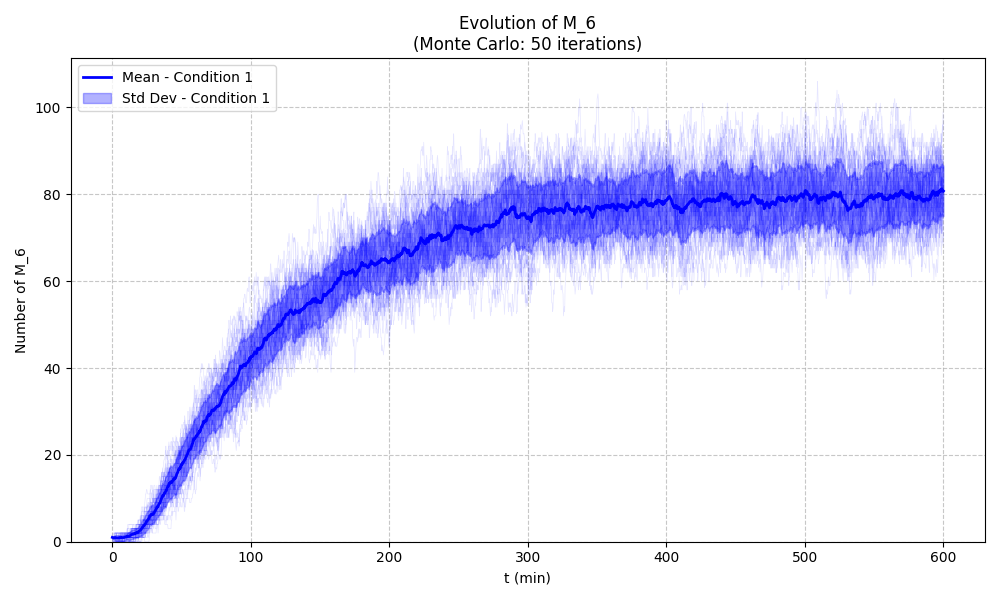}
        \caption{Simulation result from our approach}
        \label{fig:nature-evolution-m6}
    \end{subfigure}
    \hfill
    \begin{subfigure}{0.499 \textwidth}
        \centering
        \includegraphics[width=\textwidth]{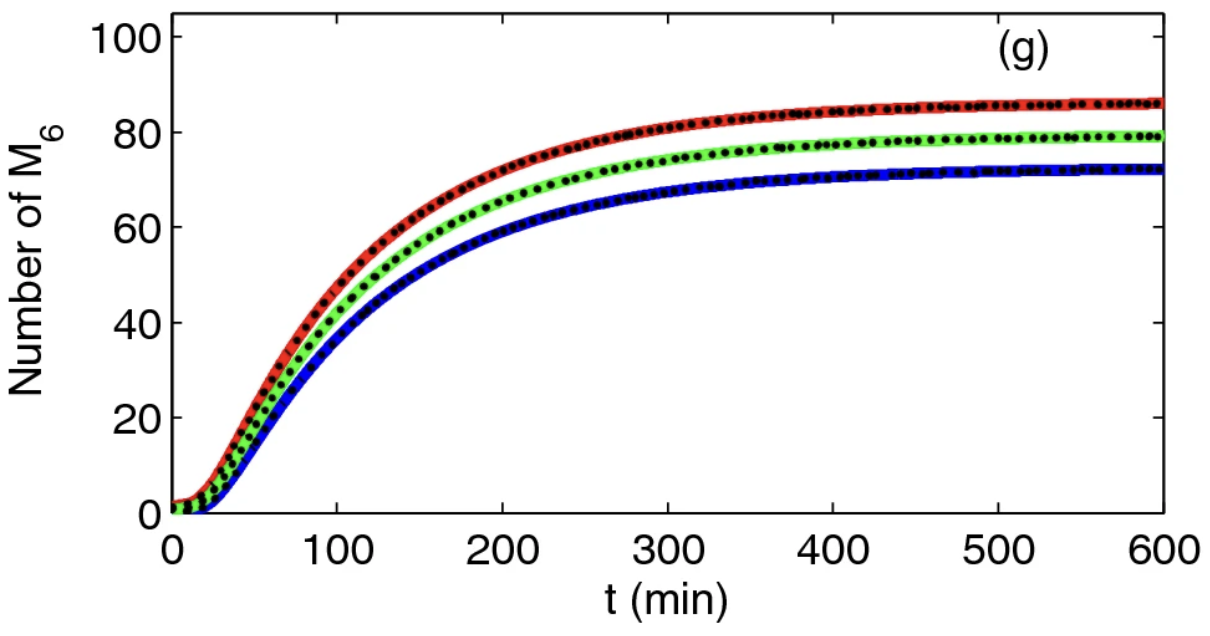}
        \caption{Original results reported in \cite{nature-oligomers}}
        \label{fig:paper-72b}
    \end{subfigure}
    \vspace{0.5cm}
    \caption{Comparison of the results from our method with the system described in \cite{nature-oligomers}, as shown in Figure 2 of \cite{nature-oligomers}}
    \label{fig:paper-72-results}
\end{figure}

\begin{figure}[H]
    \begin{subfigure}{0.48\textwidth}
        \centering
        \includegraphics[width=\textwidth, height=5cm]{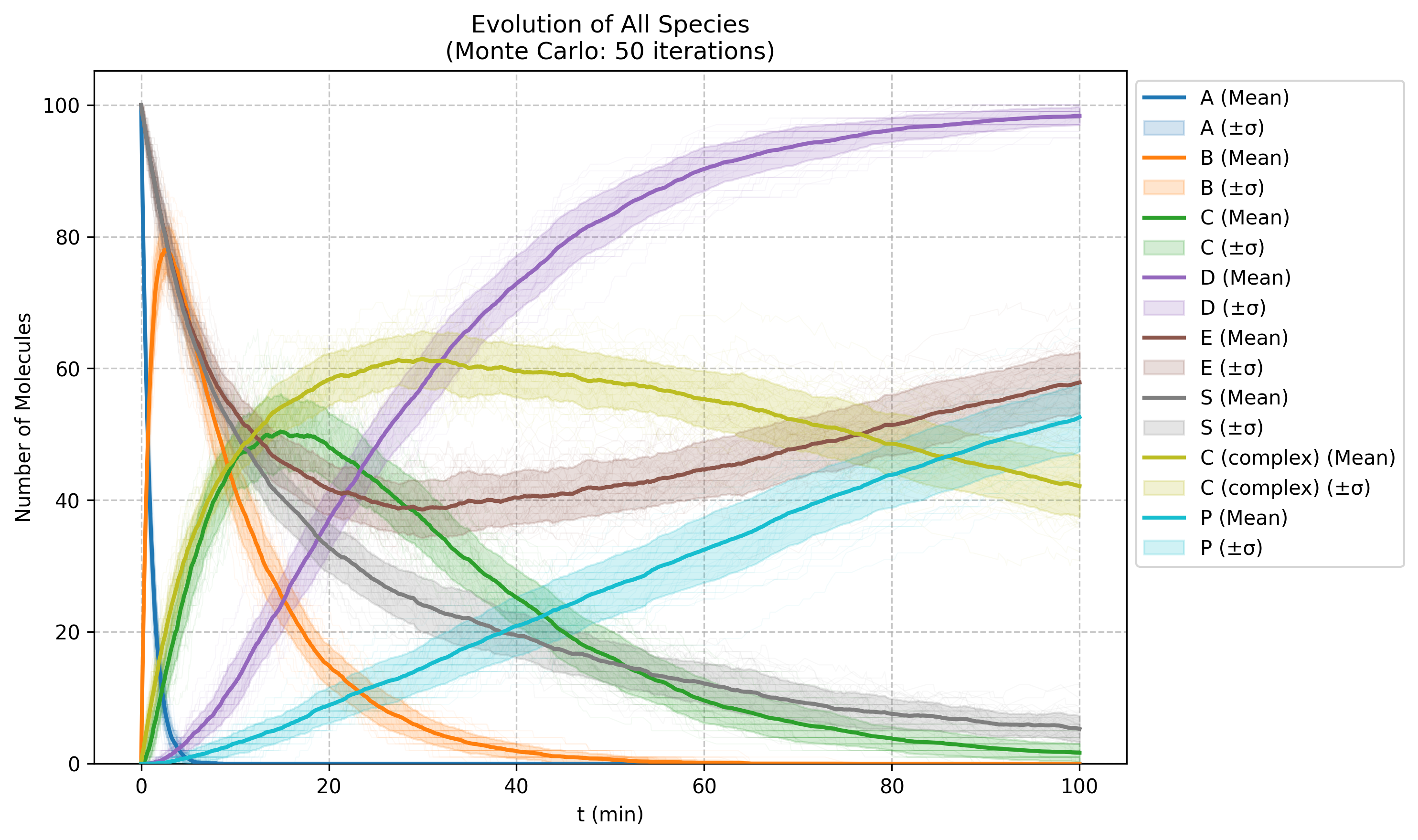}
        \caption{Results of all species from our method}
    \end{subfigure}
    \hspace*{\fill} 
    \begin{subfigure}{0.43\textwidth}
        \includegraphics[width=\textwidth, height=5cm]{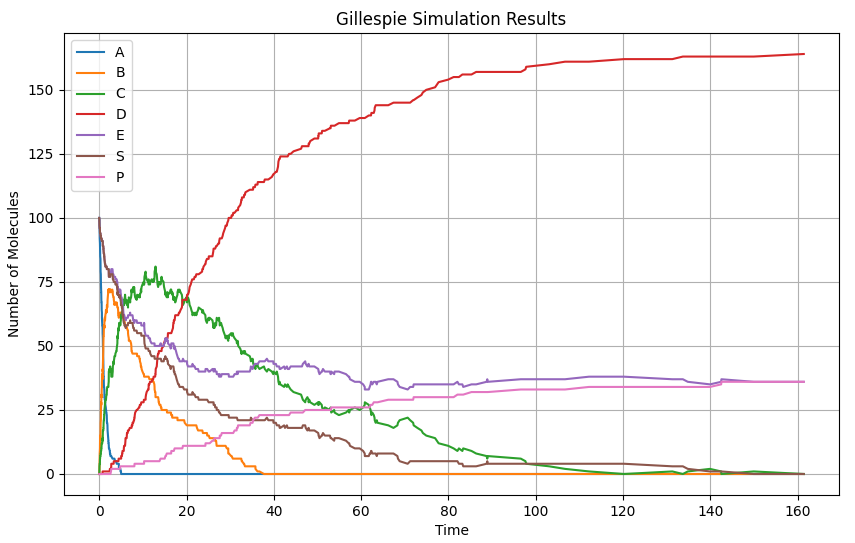}
        \caption{Results from Copasi model that our method produced}
    \end{subfigure}

    \vspace{0.5cm} 

    \begin{subfigure}{1.0\textwidth}
        \centering
        \begin{subfigure}{0.43\textwidth}
            \includegraphics[width=\textwidth, height=4.5cm]{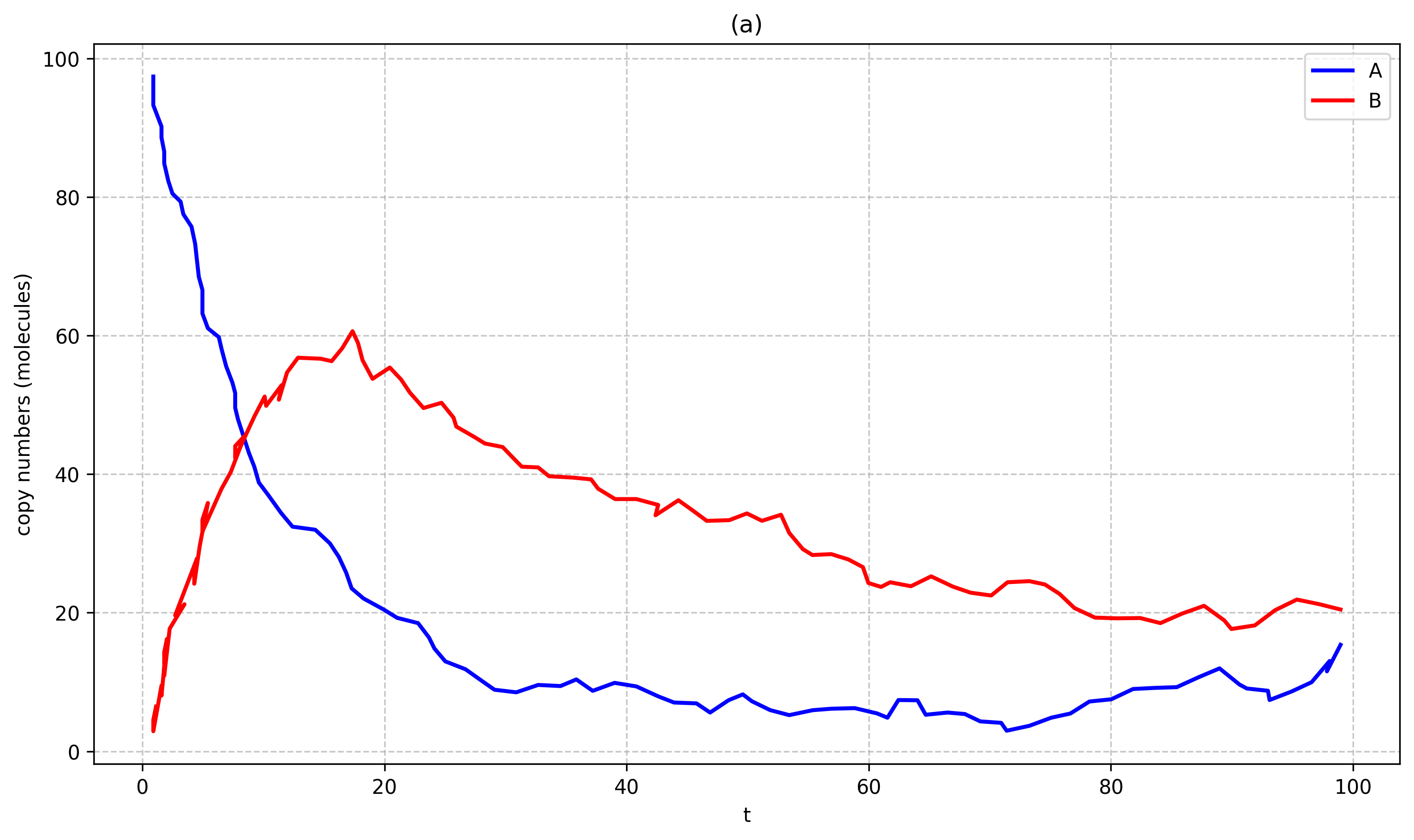}
            \caption{Original result of species 1}
        \end{subfigure}
        \hfill 
        \begin{subfigure}{0.43\textwidth}
            \includegraphics[width=\textwidth, height=4.5cm]{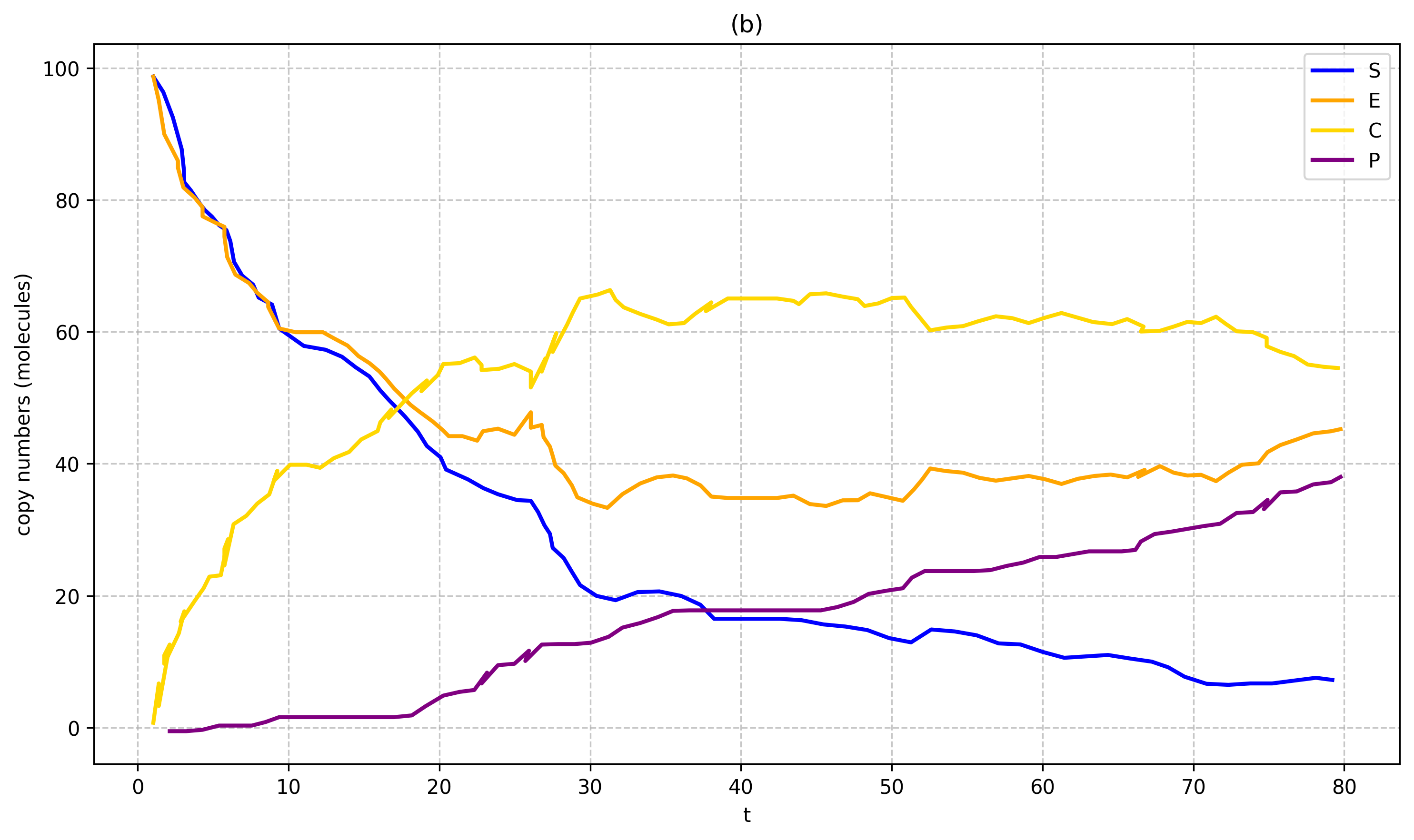}
            \caption{Original result of species 2}
        \end{subfigure}
        \label{fig:species-original}
    \end{subfigure}

    \vspace{0.5cm}
    \caption{Comparison of the results from our method with the system described in \cite{biochemical-reaction-networks}, as shown in Figure 1}
    \label{fig:paper-131-results}
\end{figure}

We perform multiple different known simulations from the literature and pass it through our approach, which includes the studies from protein aggregations \cite{Ding2024, nature-oligomers, our-1} and chemical reaction networks studies \cite{biochemical-reaction-networks}, and compare them with the results that we get by following our proposed approach. Figures 
\ref{fig:paper-45-results}, \ref{fig:paper-72-results} and \ref{fig:paper-131-results} show some results along with the results obtained from their respective original studies. More results and other relevant details are also provided in the appendix. 

\begin{figure}[H]
	\begin{center}
		\includegraphics[width=0.6\textwidth]{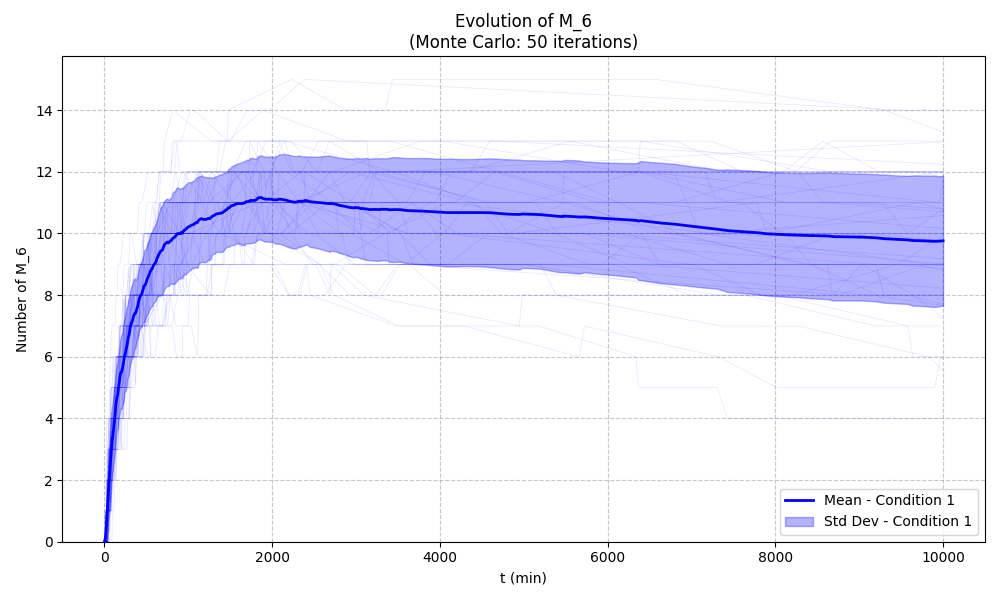}
	\end{center}
	\caption{Results plotted by analyzer step for single species evolution during the chemical reaction process}
	\label{fig:evolution-m6}
\end{figure}

\begin{figure}[H]
	\begin{center}
		\includegraphics[width=0.6\textwidth]{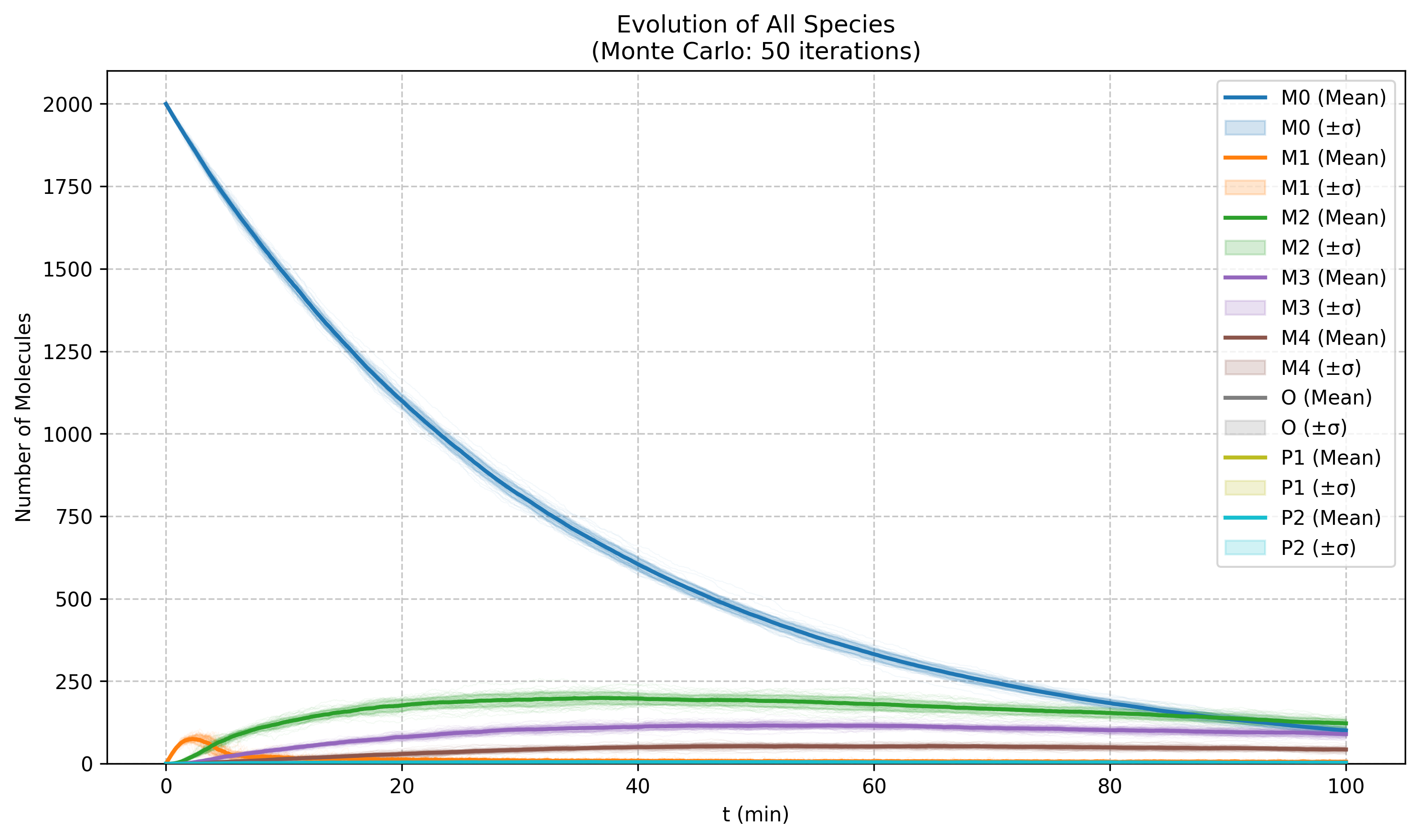}
	\end{center}
	\caption{Results obtained from the approached methods consisting 54 different reactions, comparable with the original results reported in \cite{our-1}}
	\label{fig:evolution-ad-reaction}
\end{figure}

\vspace{-0.5cm}

\vspace{-0.5cm}

\begin{figure}[H]
	\begin{center}
		\includegraphics[width=0.7\textwidth]{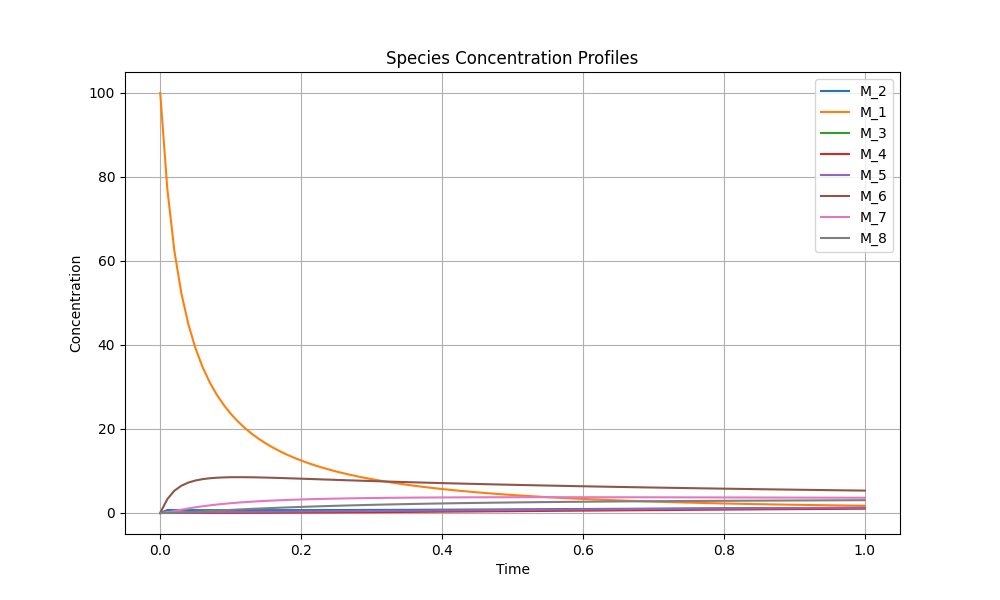}
	\end{center}
	\caption{Results obtained from integration of Copasi into the existing workflow, reaction kinetics used are same as in Figure \ref{fig:evolution-ad-reaction}}
	\label{fig:copasi-integration}
\end{figure}

 Figure \ref{fig:evolution-m6} and \ref{fig:evolution-ad-reaction} show the results obtained from our approach where the dynamics of both single and multiple species are simulated. The species, reaction kinetics and stochiometry matrix were identified as shown in the appendix \ref{tab:chemical-species}, \ref{tab:chemical-reactions} and \ref{tab:stoichiometry}. The monte carlo simulation was run for number of identified iterations and the results were aggregated and plotted by the last step. Figure \ref{fig:copasi-integration} was produced by the automatic integration of Copasi software \cite{hoops2006copasi} in the existing workflow without any intervention from the user. The parsed chemical reactions are automatically modeled into Copasi and produced model can be directly imported in Copasi software as required.   

To evaluate the efficiency of our method, we perform number of comprehensive simulations on the published studies. Figure \ref{fig:paper-45-results}, \ref{fig:paper-72-results} and \ref{fig:paper-131-results} show a comparison between the original results and the dynamics obtained using our method, based on the reaction descriptions provided in the respective studies. As evident from these figures, our method accurately simulates the dynamics without the need for manual preparation of reaction kinetics.

\section{Integration with Copasi}
After the species are parsed, the reaction kinetics and needed parameters are identified. We then leveraged the Copasi's python integration (basico) \cite{basico} and directly build Copasi model without any intervention of the user. This gives the researchers and modelers power to save hours, and building the Copasi model manually by inserting each species and reactions into the software, defining the tasks and reporting criteria. We not only created stochastic model, but also created deterministic model, which can be directly simulated inside the Copasi. The result we get from this integration is identical with the ones we retrieved from the monte carlo simulation and the ones reported in the original studies. 

\section{Integration with Open Source LLM}
We also tested this approach with open source LLM, such as llama-3.1 8B parameters model, and we found that it can extract all the species and reaction kinetics for smaller reaction descriptions. But if we increase the number of reactions, then it fails to parse the details of the reaction correctly, thus failing to go to the next step. Even though, the smaller model may not parse longer details due to their inherent limitations, larger open source models may be able to tackle this challenge. 

\begin{figure}[H]
	\begin{center}
		\includegraphics[width=0.8\textwidth]{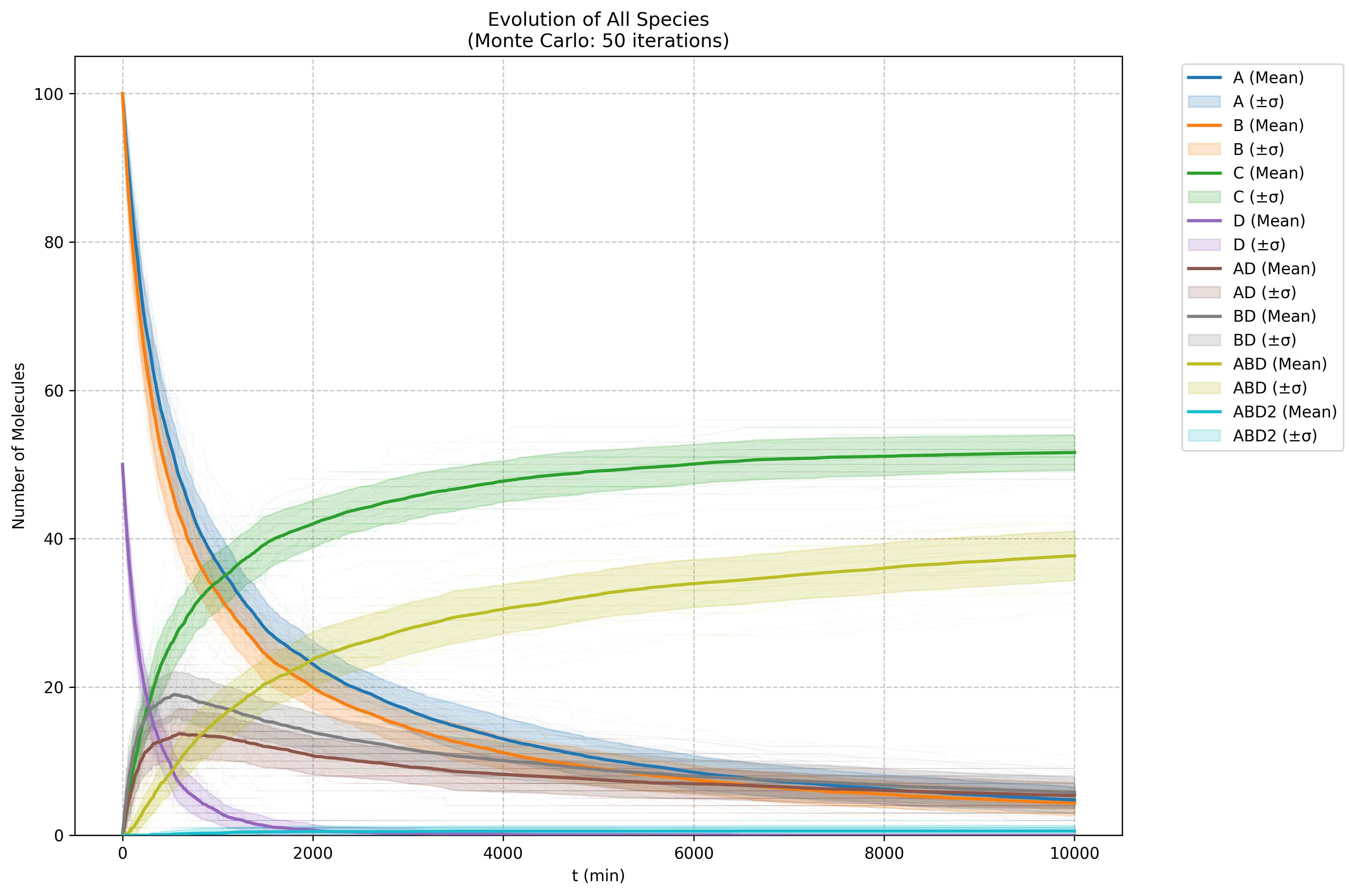}
	\end{center}
	\caption{Results obtained for 4 chemical species using open source LLM llama-3.1 8B}
	\label{fig:ollama-results}
\end{figure}
Figure \ref{fig:ollama-results} shows a simple stochastic simulation of chemical reaction network with 4 species and all combination of their reactions and products. This also shows the promise of open source large language models in complex task like parsing chemical reactions from input texts.

\section{Conclusion}
In this study, we leveraged popular large language models such as OpenAI GPT, Claude and LLMA to parse the chemical reactions from textual description, then converted them into reaction kinetics, stochiometry matrix and ultimately created the stochastic simulation. We also integrated the overall process with Copasi, thus easing the process of building Copasi model. With Copasi integration, one can do both deterministic as well as stochastic simulation. Our work shows how large language models can be used to automate some of the manual tasks which takes considerable time and effort during the modelling and simulation processes.

\section*{Acknowledgements}
We would like to extend our gratitude to our colleagues who read this manuscript and provided valuable feedback.







\pagebreak
\section*{Appendix}

\subsection*{Example of Chemical Species}

\begin{table}[H]
    \centering
    \begin{tabular}{|l|c|}
        \toprule
        \textbf{Name} & \textbf{Initial Amount} \\
        \midrule
        A      & 100.0 \\
        B      & 0.0 \\
        C\_mono & 0.0 \\
        D      & 0.0 \\
        E      & 100.0 \\
        S      & 100.0 \\
        ES     & 0.0 \\
        P      & 0.0 \\
        \bottomrule
    \end{tabular}
    \vspace{0.5cm}
    \caption{Initial Amounts of Chemical Species}
    \label{tab:chemical-species}
\end{table}

\subsection*{Example of Chemical Reactions}

\begin{table}[H]
    \centering
    \begin{tabular}{|l|c|c|c|}
        \toprule
        \textbf{Name} & \textbf{Reactants} & \textbf{Products} & \textbf{Rate Constant} \\
        \midrule
        mono\_chain\_r1 & 1A        & 1B       & 1.0 \\
        mono\_chain\_r2 & 1B        & 1C\_mono  & 0.1 \\
        mono\_chain\_r3 & 1C\_mono   & 1D       & 0.05 \\
        enzyme\_r1     & 1E + 1S   & 1ES      & 0.001 \\
        enzyme\_r2     & 1ES       & 1E + 1S  & 0.005 \\
        enzyme\_r3     & 1ES       & 1E + 1P  & 0.01 \\
        \bottomrule
    \end{tabular}
        \vspace{0.5cm}

    \caption{Chemical Reactions and Rate Constants}
    \label{tab:chemical-reactions}
\end{table}

\subsection*{Example of Stoichiometry Matrix} 

\begin{table}[H]
    \centering
    \renewcommand{\arraystretch}{1.2}
    \begin{tabular}{c|cccccc}
        & \text{mono\_chain\_r1} & \text{mono\_chain\_r2} & \text{mono\_chain\_r3} & \text{enzyme\_r1} & \text{enzyme\_r2} & \text{enzyme\_r3} \\\hline
        A    & -1.0 &  0.0 &  0.0 &  0.0 &  0.0 &  0.0 \\
        B    &  1.0 & -1.0 &  0.0 &  0.0 &  0.0 &  0.0 \\
       C\_mono &  0.0 &  1.0 & -1.0 &  0.0 &  0.0 &  0.0 \\
        D    &  0.0 &  0.0 &  1.0 &  0.0 &  0.0 &  0.0 \\
        E    &  0.0 &  0.0 &  0.0 & -1.0 &  1.0 &  1.0 \\
        S    &  0.0 &  0.0 &  0.0 & -1.0 &  1.0 &  0.0 \\
        ES   &  0.0 &  0.0 &  0.0 &  1.0 & -1.0 & -1.0 \\
        P    &  0.0 &  0.0 &  0.0 &  0.0 &  0.0 &  1.0 \\
    \end{tabular}
    \vspace{0.5cm}
    \caption{Stoichiometry matrix representing the reactions in the system}
    \label{tab:stoichiometry}
\end{table}


\subsection*{Example Input for the system described in Paper \cite{Ding2024}}
This system models a misfolding-driven aggregation process, where monomers undergo misfolding, oligomerization, and dissociation. Some reactions are set to zero, indicating they are inactive or negligible.

Reaction Pathways and Rate Constants

Misfolding Reaction \\
 Spontaneous misfolding: A normal monomer misfolds into an intermediate state.\\  
 \( M_0 \rightarrow M_1 \) with rate constant \( K_0^+ = 0.01 \).\\

Primary Nucleation  \\
 Formation of the first stable oligomer: Two misfolded monomers combine.\\  
   \( M_1 + M_1 \rightarrow M_2 \) with rate constant \( K_1^+ = 3.8 \times 10^{-3} \).\\

Oligomer Formation  \\
 Stepwise growth of oligomers:  \\
   \( M_1 + M_2 \rightarrow M_3 \) with rate constant \( K_2^+ = 4.8 \times 10^{-3} \).  \\
   \( M_1 + M_3 \rightarrow M_4 \) with rate constant \( K_3^+ = 5.8 \times 10^{-3} \).  \\
   \( M_1 + M_4 \rightarrow O \) with rate constant \( K_4^+ = 6.8 \times 10^{-3} \).\\

Dissociation Reactions  \\
 Breakdown of oligomers back into smaller components:  \\
   \( M_2 \rightarrow M_1 + M_1 \) with rate constant \( K_d = 0.36 \).  \\
   \( M_3 \rightarrow M_1 + M_2 \) with rate constant \( K_d = 0.36 \).  \\
   \( M_4 \rightarrow M_1 + M_3 \) with rate constant \( K_d = 0.36 \).  \\
   \( O \rightarrow M_1 + M_4 \) with rate constant \( K_d = 0.36 \).\\

Inactive or Negligible Reactions (Rate Constant = 0)  \\
These reactions do not occur in this system, as their rate constants are set to zero:  \\
 Secondary nucleation: \( M_2 + M_2 \rightarrow M_3 \), \( K_{se} = 0 \).\\  
 Catalytic conversion: \( O \rightarrow M_0 \), \( K_c = 0 \).  \\
 General polymerization: \( M_1 + M_4 \rightarrow P \), \( K_+ = 0 \).\\

Additional Parameters \\
 Oligomer Size Thresholds:  \\
   Minimum size for oligomer formation: \( O_\alpha = 6 \).  \\
   Threshold for polymerization: \( P_\alpha = 10 \).\\

 Degradation Rate: \( \delta = 0 \) (No degradation).  \\
 System Size Factor: \( \gamma = 4000 \) (Controls reaction scaling).  \\

Initial Conditions  \\
At the beginning of the reaction:  \\
 Monomers (unfolded state): \( N_{M_0}(0) = 2000 \).  \\
 All other species start at zero:  \\
  \( N_{M_1}(0) = N_{M_2}(0) = N_{M_3}(0) = N_{M_4}(0) = N_O(0) = 0 \).\\

\begin{figure}[H]
    \centering

    \begin{subfigure}{1.0\textwidth}
        \centering
        \includegraphics[width=\textwidth]{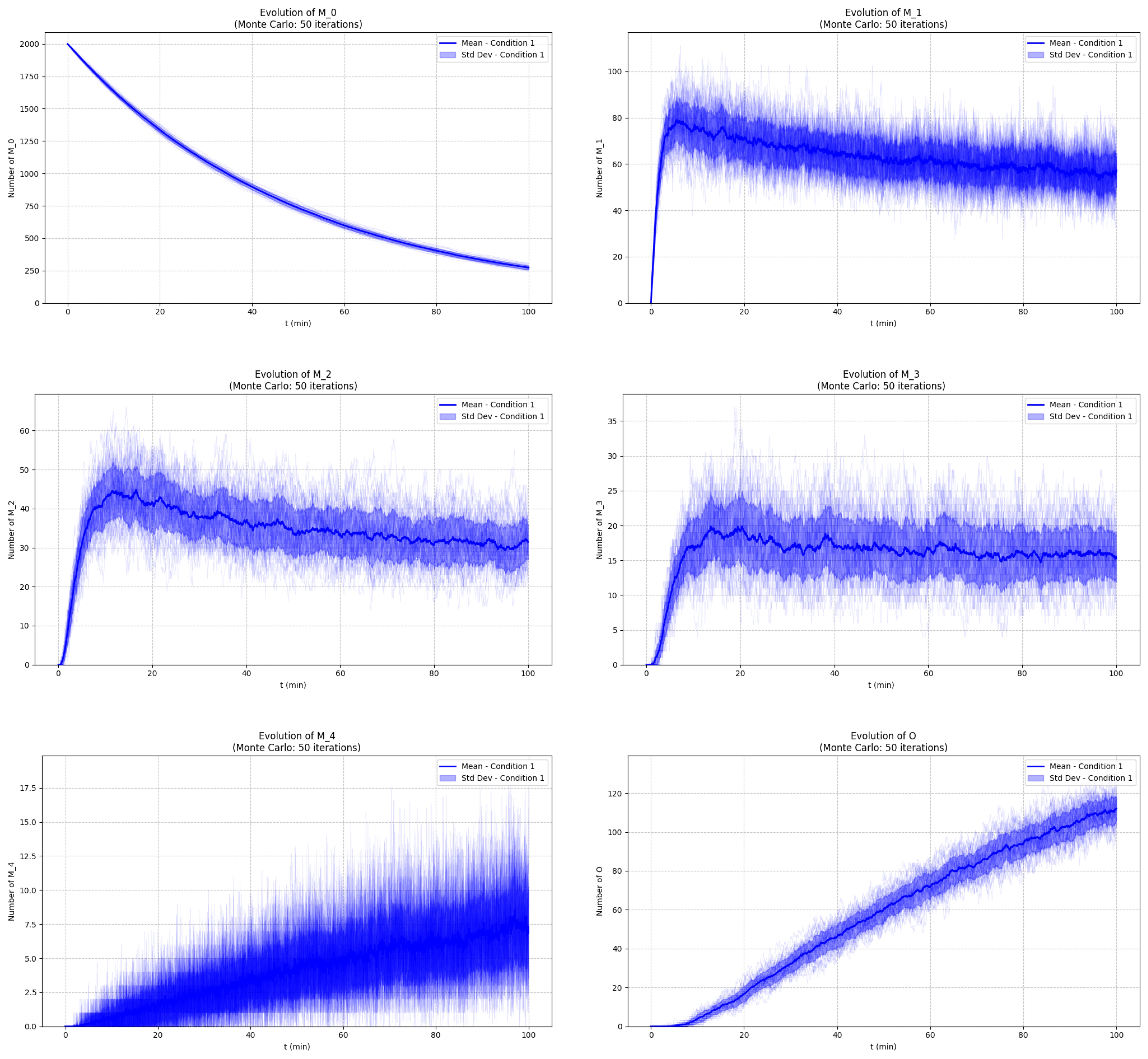}
        \caption{Results of each species}
        \label{fig:paper45-collage}
    \end{subfigure}

    \vspace{0.5cm} 



    \begin{subfigure}{0.48\textwidth}
        \centering
        \includegraphics[width=\textwidth]{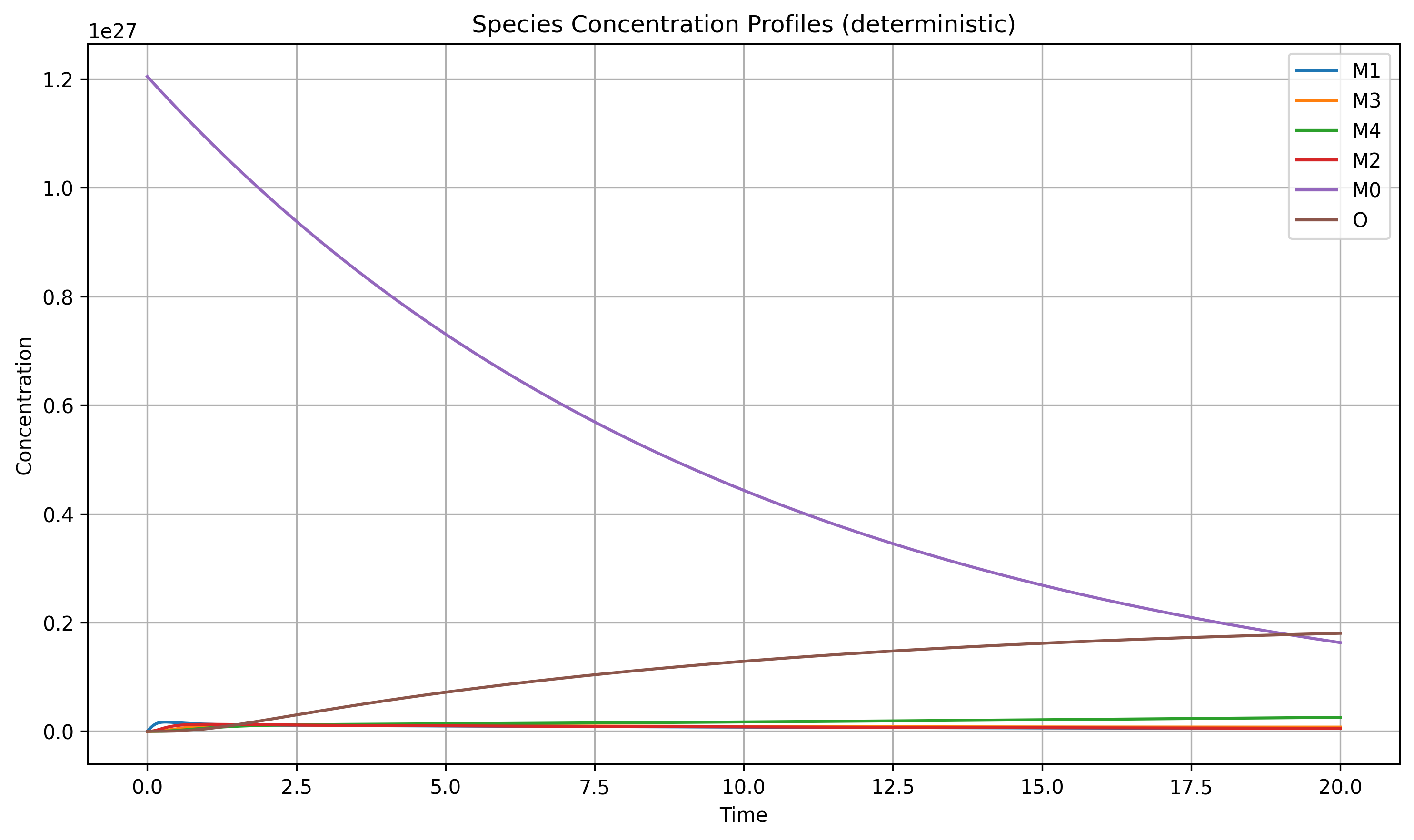}
        \caption{Result of Copasi (Deterministic)}
        \label{fig:copasi-deterministic}
    \end{subfigure}
    \hfill
    \begin{subfigure}{0.48\textwidth}
        \centering
        \includegraphics[width=\textwidth]{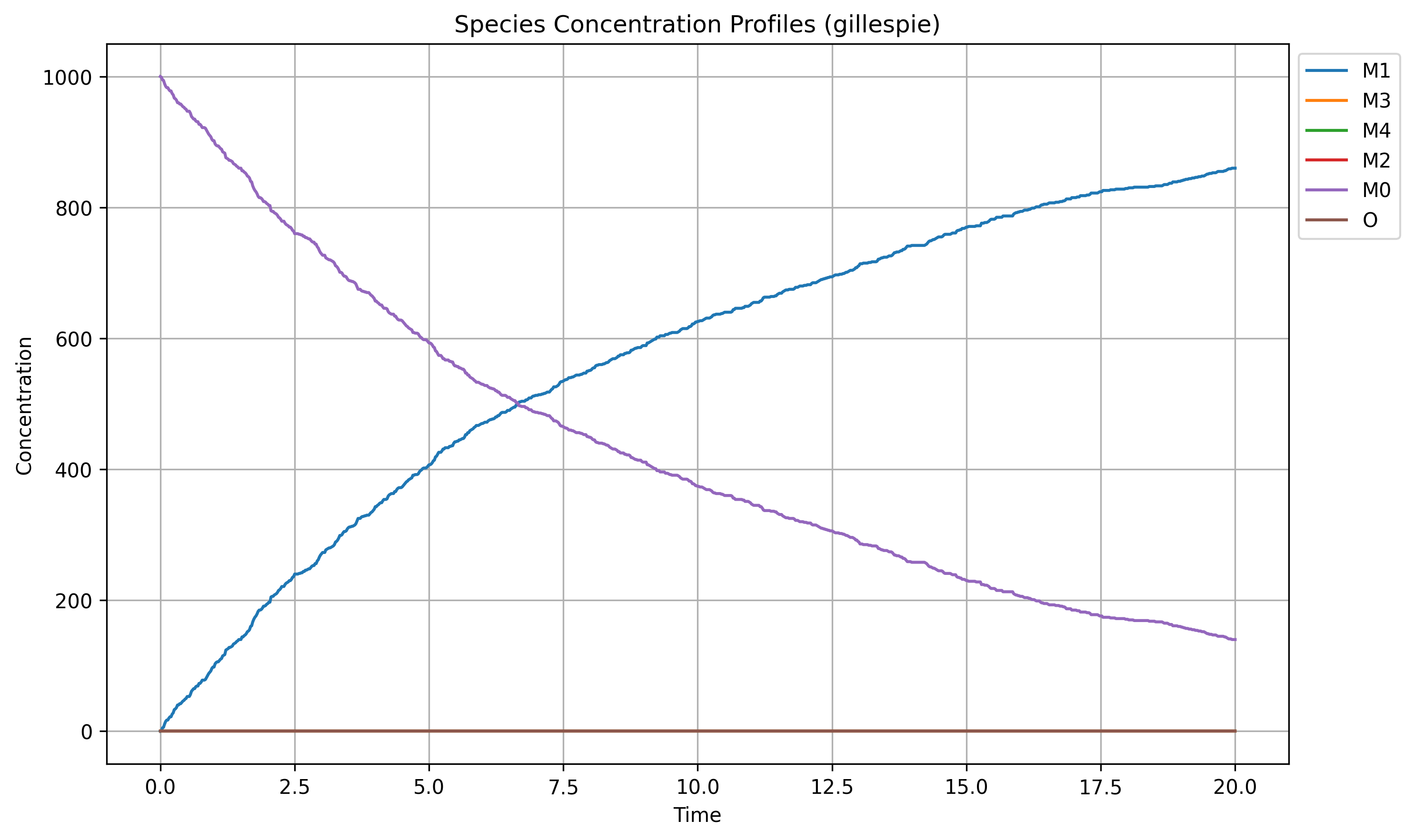}
        \caption{Result of Copasi (Stochastic Gillespie)}
        \label{fig:copasi-gillespie}
    \end{subfigure}

    \vspace{0.5cm}

    \caption{Results from our method for the system described in Paper \cite{Ding2024}.}
    \label{fig:all-results}
\end{figure}

\subsection*{Example Input for the system described in Paper \cite{nature-oligomers}}

Consider a complex reaction system: \\
Reactions and Rate Constants \\

Monomer production \\  
\( \emptyset \rightarrow M_0 \) with rate constant \( k_p = 0 \) (no monomer production). \\

Misfolding \\  
\( M_0 \rightarrow M_1 \) with rate constant \( K_0 = 0.01 \) min\(^{-1}\). \\

Aggregation \\  
\( M_1 + M_1 \rightarrow M_2 \) with rate constant \( K_a = 0.002 \) min\(^{-1}\). \\  
\( M_1 + M_2 \rightarrow M_3 \) with rate constant \( K_a = 0.002 \) min\(^{-1}\). \\  
\( M_1 + M_3 \rightarrow M_4 \) with rate constant \( K_a = 0.002 \) min\(^{-1}\). \\  
\( M_1 + M_4 \rightarrow M_5 \) with rate constant \( K_a = 0.002 \) min\(^{-1}\). \\  
\( M_1 + M_5 \rightarrow M_6 \) with rate constant \( K_a = 0.002 \) min\(^{-1}\). \\

Dissociation \\  
\( M_2 \rightarrow M_1 + M_1 \) with rate constant \( K_b = 0.1 \) min\(^{-1}\). \\  
\( M_3 \rightarrow M_1 + M_2 \) with rate constant \( K_b = 0.1 \) min\(^{-1}\). \\  
\( M_4 \rightarrow M_1 + M_3 \) with rate constant \( K_b = 0.1 \) min\(^{-1}\). \\  
\( M_5 \rightarrow M_1 + M_4 \) with rate constant \( K_b = 0.1 \) min\(^{-1}\). \\  
\( M_6 \rightarrow M_1 + M_5 \) with rate constant \( K_6 = 0.1 \) min\(^{-1}\). \\

Initial Concentrations \\  
\( M_0 \) (monomers): \( N_{M_0}(0) = 2000 \). \\  
Other species: \( N_{M_1}(0) = N_{M_2}(0) = N_{M_3}(0) = N_{M_4}(0) = N_{M_5}(0) = N_{M_6}(0) = 1 \). \\

\begin{figure}[H]
    \centering
    \includegraphics[width=\textwidth]{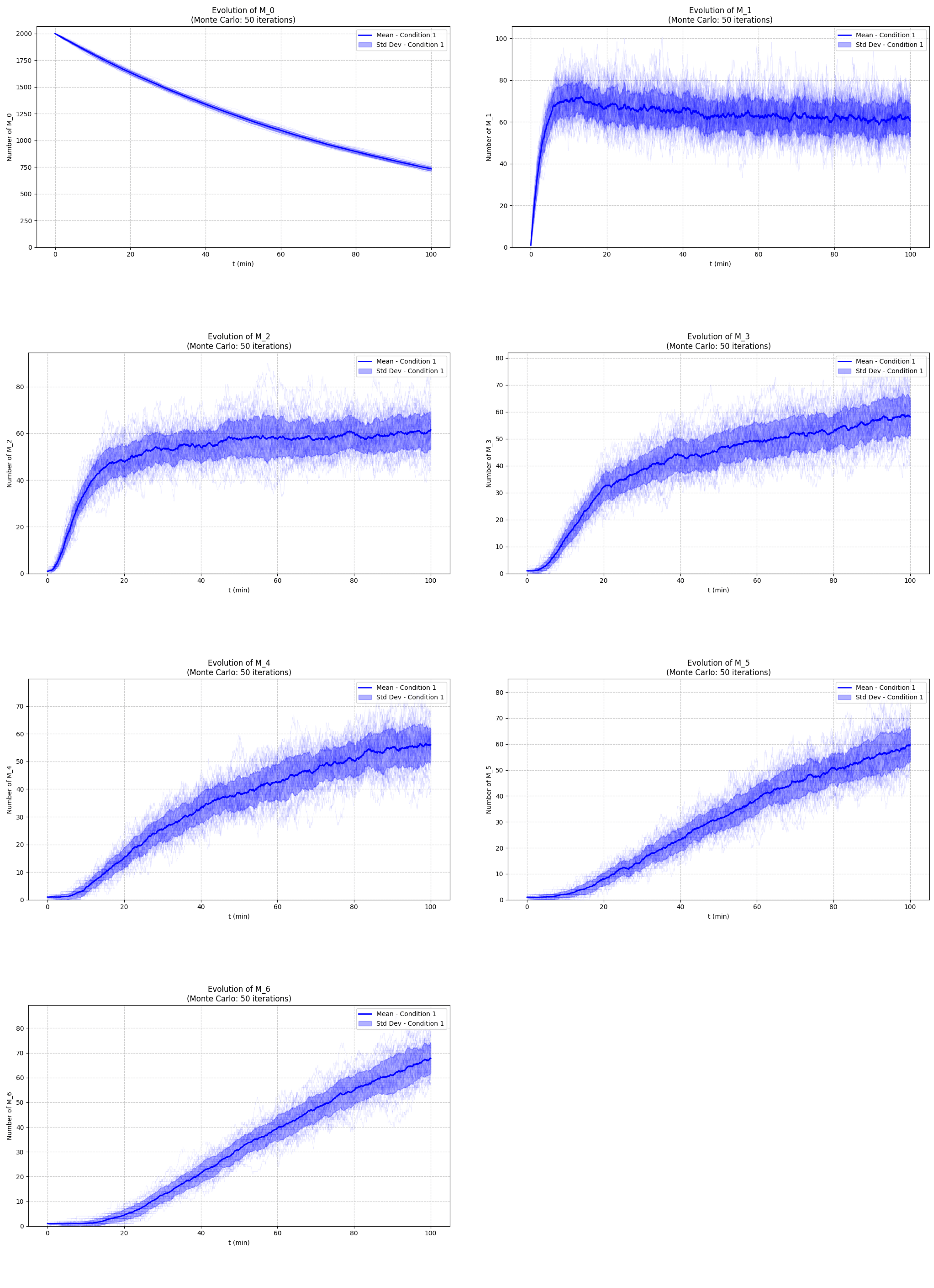}
    \caption{Results of each species}
    \label{fig:nature-collage}
\end{figure}

\begin{figure}[H]
    \centering
    \begin{subfigure}{0.48\textwidth}
        \centering
        \includegraphics[width=\textwidth]{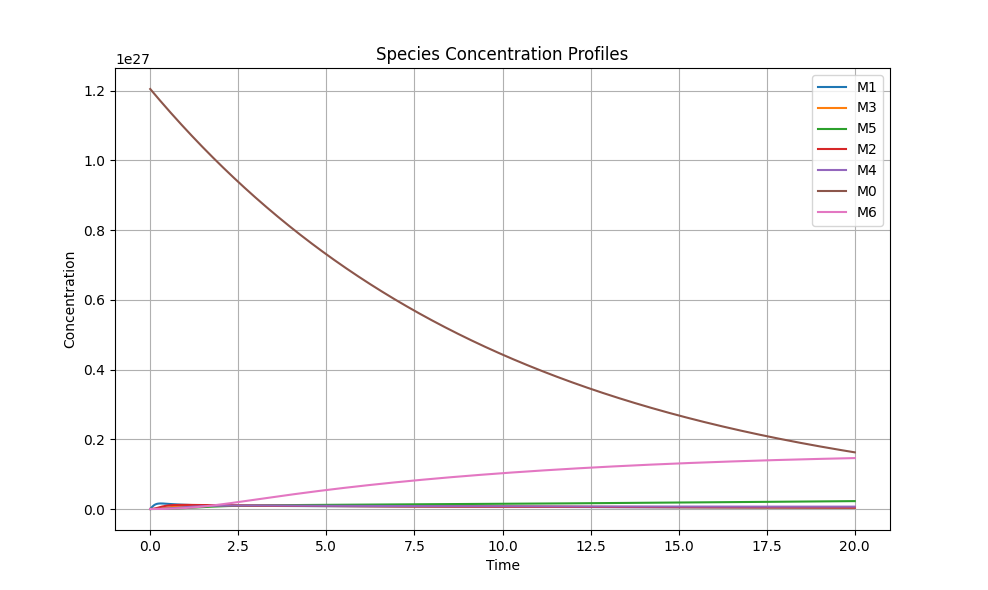}
        \caption{Result from Copasi (Deterministic)}
        \label{fig:nature-copasi-deterministic}
    \end{subfigure}
    \hfill
    \begin{subfigure}{0.48\textwidth}
        \centering
        \includegraphics[width=\textwidth]{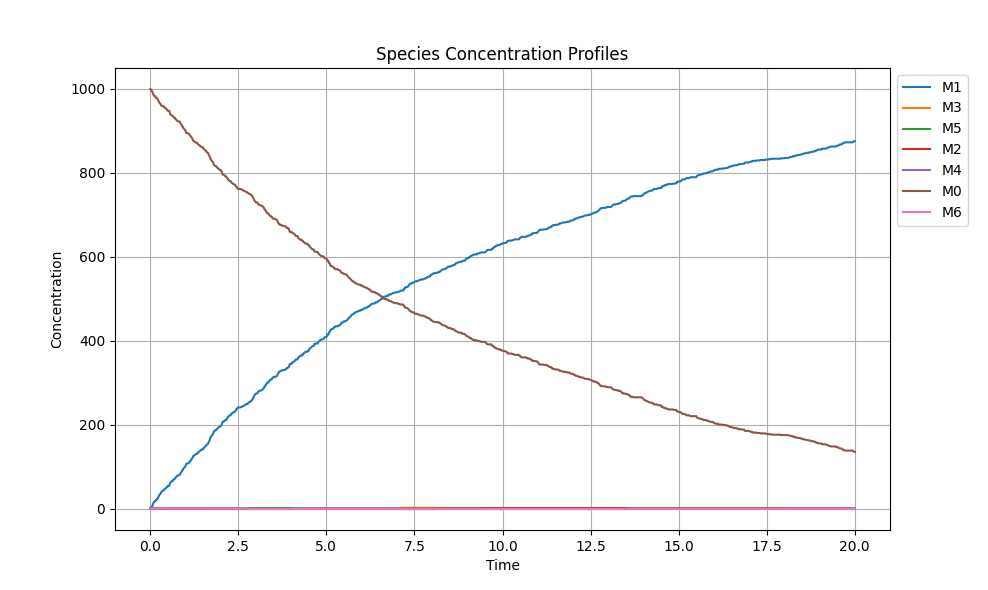}
        \caption{Result from Copasi (Stochastic Gillespie)}
        \label{fig:nature-copasi-gillespie}
    \end{subfigure}
    
    \caption{Results from Copasi simulations for the system described in Paper \cite{nature-oligomers}.}
    \label{fig:nature-copasi-results}
\end{figure}

\subsection*{Example Input for the system described in Paper \cite{our-1}}

Consider a complex reaction system: \\
Species M1 and M1 combine to form M2 with rate constant \( k_0 = 0.00001 \). \\  
Species M1 and M2 combine to form M3 with rate constant \( k_2 = 0.000009 \). \\  
Species M1 and M3 combine to form M4 with rate constant \( k_4 = 0.000008 \). \\  
Species M2 and M2 combine to form M4 with rate constant \( k_6 = 0.000007 \). \\  
Species M1 and M4 combine to form M5 with rate constant \( k_8 = 0.0000065 \). \\  
Species M1, M2, and M2 combine to form M5 with rate constant \( k_{10} = 0.000006 \). \\  
Species M2 and M3 combine to form M5 with rate constant \( k_{12} = 0.0000055 \). \\  
Species M1 and M5 combine to form M6 with rate constant \( k_{14} = 0.000005 \). \\  
Species M2 and M4 combine to form M6 with rate constant \( k_{16} = 0.0000045 \). \\  
Species M3 and M3 combine to form M6 with rate constant \( k_{18} = 0.000004 \). \\  
Species M2, M2, M1, and M1 combine to form M6 with rate constant \( k_{20} = 0.0000035 \). \\  
Species M1 and M6 combine to form M7 with rate constant \( k_{22} = 0.000003 \). \\  
Species M2 and M5 combine to form M7 with rate constant \( k_{24} = 0.0000028 \). \\  
Species M3, M3, and M1 combine to form M7 with rate constant \( k_{26} = 0.0000026 \). \\  
Species M3 and M4 combine to form M7 with rate constant \( k_{28} = 0.0000024 \). \\  
Species M1 and 3M2 combine to form M7 with rate constant \( k_{30} = 0.0000022 \). \\  
Species M1 and M7 combine to form M8 with rate constant \( k_{32} = 0.000002 \). \\  
Species M2 and M6 combine to form M8 with rate constant \( k_{34} = 0.0000018 \). \\  
Species M3 and M5 combine to form M8 with rate constant \( k_{36} = 0.0000016 \). \\  
Species M4 and M4 combine to form M8 with rate constant \( k_{38} = 0.0000014 \). \\  
Species M1, M2, and M5 combine to form M8 with rate constant \( k_{40} = 0.0000013 \). \\  
Species M1, M4, and M3 combine to form M8 with rate constant \( k_{42} = 0.0000012 \). \\  
Species M2, M2, and M4 combine to form M8 with rate constant \( k_{44} = 0.0000011 \). \\  
Species M2, M3, and M3 combine to form M8 with rate constant \( k_{46} = 0.000001 \). \\  
Species M1, M1, M3, and M3 combine to form M8 with rate constant \( k_{48} = 0.0000009 \). \\  
Species 4M2 combine to form M8 with rate constant \( k_{50} = 0.0000008 \). \\  
Species M1, M1, M2, and M4 combine to form M8 with rate constant \( k_{52} = 0.0000007 \). \\  

Decomposition Reactions and Rate Constants \\
Species M2 decomposes into 2M1 with rate constant \( k_1 = 0.000009 \). \\  
Species M3 decomposes into M1 and M2 with rate constant \( k_3 = 0.000008 \). \\  
Species M4 decomposes into M1 and M3 with rate constant \( k_5 = 0.000007 \). \\  
Species M4 decomposes into 2M2 with rate constant \( k_7 = 0.0000065 \). \\  
Species M5 decomposes into M1 and M4 with rate constant \( k_9 = 0.000006 \). \\  
Species M5 decomposes into M1 and 2M2 with rate constant \( k_{11} = 0.0000055 \). \\  
Species M5 decomposes into M2 and M3 with rate constant \( k_{13} = 0.000005 \). \\  
Species M6 decomposes into M1 and M5 with rate constant \( k_{15} = 0.0000045 \). \\  
Species M6 decomposes into M2 and M4 with rate constant \( k_{17} = 0.000004 \). \\  
Species M6 decomposes into 2M3 with rate constant \( k_{19} = 0.0000035 \). \\  
Species M6 decomposes into 2M2 and 2M1 with rate constant \( k_{21} = 0.000003 \). \\  
Species M7 decomposes into M1 and M6 with rate constant \( k_{23} = 0.0000028 \). \\  
Species M7 decomposes into M2 and M5 with rate constant \( k_{25} = 0.0000026 \). \\  
Species M7 decomposes into M3, M3, and M1 with rate constant \( k_{27} = 0.0000024 \). \\  
Species M7 decomposes into M3 and M4 with rate constant \( k_{29} = 0.0000022 \). \\  
Species M7 decomposes into M1 and 3M2 with rate constant \( k_{31} = 0.000002 \). \\  
Species M8 decomposes into M1 and M7 with rate constant \( k_{33} = 0.0000018 \). \\  
Species M8 decomposes into M2 and M6 with rate constant \( k_{35} = 0.0000016 \). \\  
Species M8 decomposes into M3 and M5 with rate constant \( k_{37} = 0.0000014 \). \\  
Species M8 decomposes into 2M4 with rate constant \( k_{39} = 0.0000013 \). \\  
Species M8 decomposes into M1, M2, and M5 with rate constant \( k_{41} = 0.0000012 \). \\  
Species M8 decomposes into M1, M3, and M4 with rate constant \( k_{43} = 0.0000011 \). \\  
Species M8 decomposes into M2 and 2M3 with rate constant \( k_{45} = 0.000001 \). \\  
Species M8 decomposes into 4M2 with rate constant \( k_{51} = 0.0000008 \). \\  
Species M8 decomposes into 2M1, M2, and M4 with rate constant \( k_{53} = 0.0000007 \). \\  

Initial Concentrations \\
\( M_1 = 10000 \) \\  
All other species = 1 \\  

\begin{figure}[H]
    \centering
    \includegraphics[width=\textwidth]{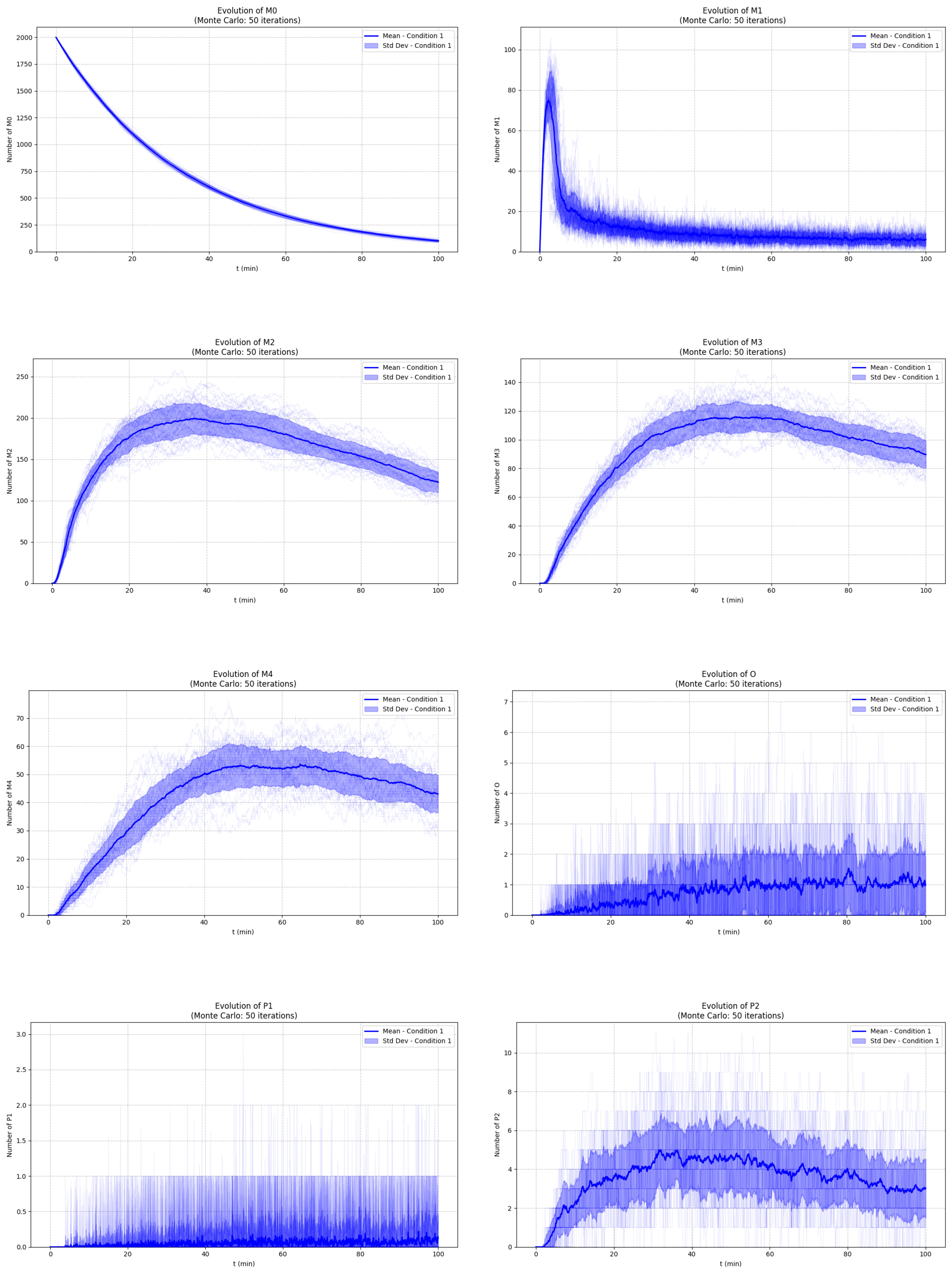}
    \caption{Results of each species}
    \label{fig:paper10-collage}
\end{figure}

\begin{figure}[H]
    \centering
    \includegraphics[width=0.55\textwidth]{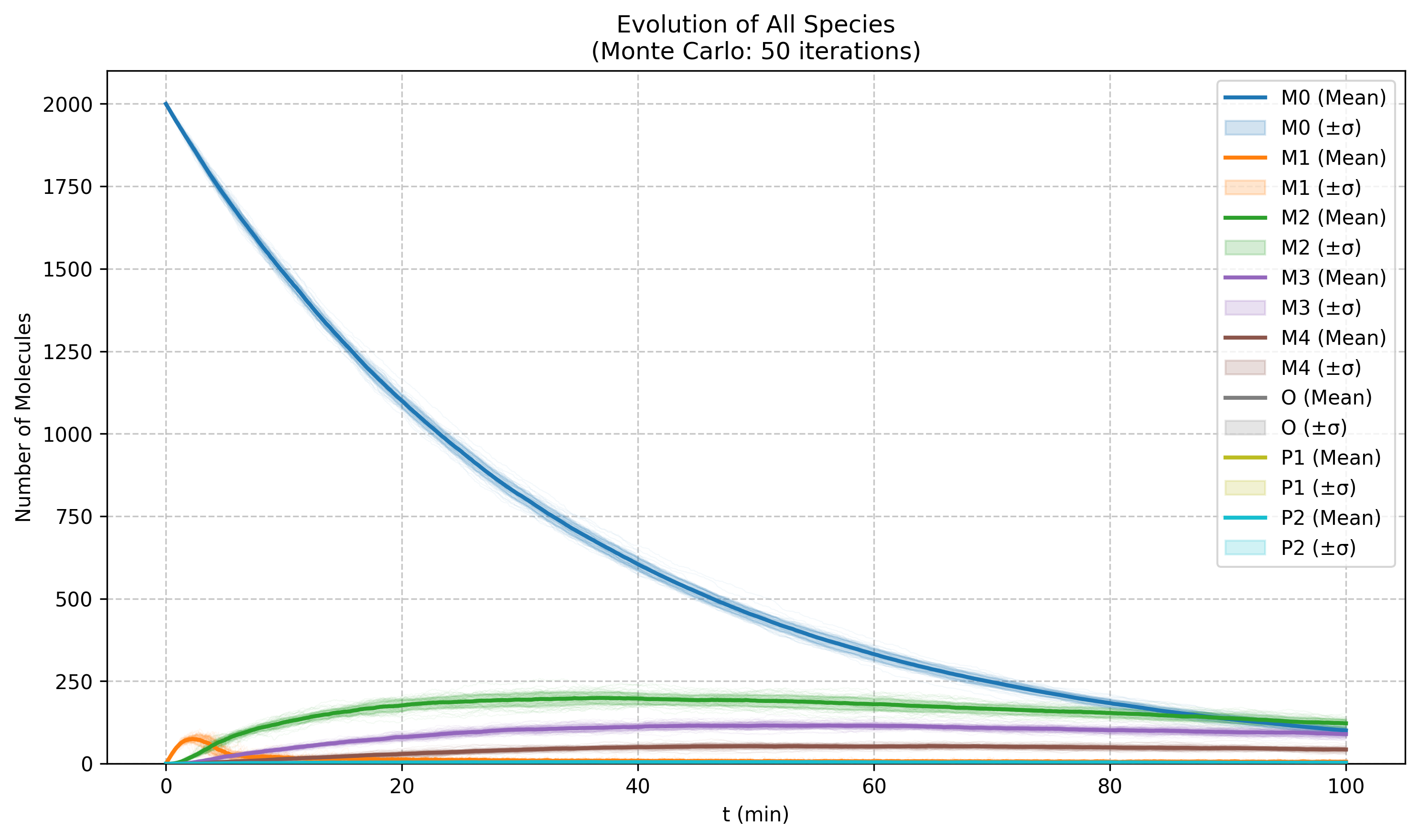}
    \caption{Result of all species}
    \label{fig:paper10-evolution-all-species}
\end{figure}


\subsection*{Example Input for the system described in Paper \cite{biochemical-reaction-networks}}

Biochemical Reaction System Description \\

Mono-Molecular Chain Model \\  
Reaction 1: Species A transforms into species B with rate constant k1 = 1.0. \\  
Reaction 2: Species B transforms into species C with rate constant k2 = 0.1. \\  
Reaction 3: Species C transforms into species D with rate constant k3 = 0.05. \\  

Enzyme Kinetics Model \\  
Reaction 1: Enzyme E and substrate S form an enzyme-substrate complex C with rate constant k1 = 0.001. \\  
Reaction 2: The enzyme-substrate complex C dissociates back into E and S with rate constant k2 = 0.005. \\  
Reaction 3: The enzyme-substrate complex C converts into enzyme E and product P with rate constant k3 = 0.01. \\  

Initial Concentrations \\  
Mono-Molecular Chain Model \\  
A = 100, B = 0, C = 0, D = 0. \\  

Enzyme Kinetics Model \\  
E = 100, S = 100, C = 0, P = 0. \\  

\begin{figure}[H]
    \centering
    \includegraphics[width=\textwidth]{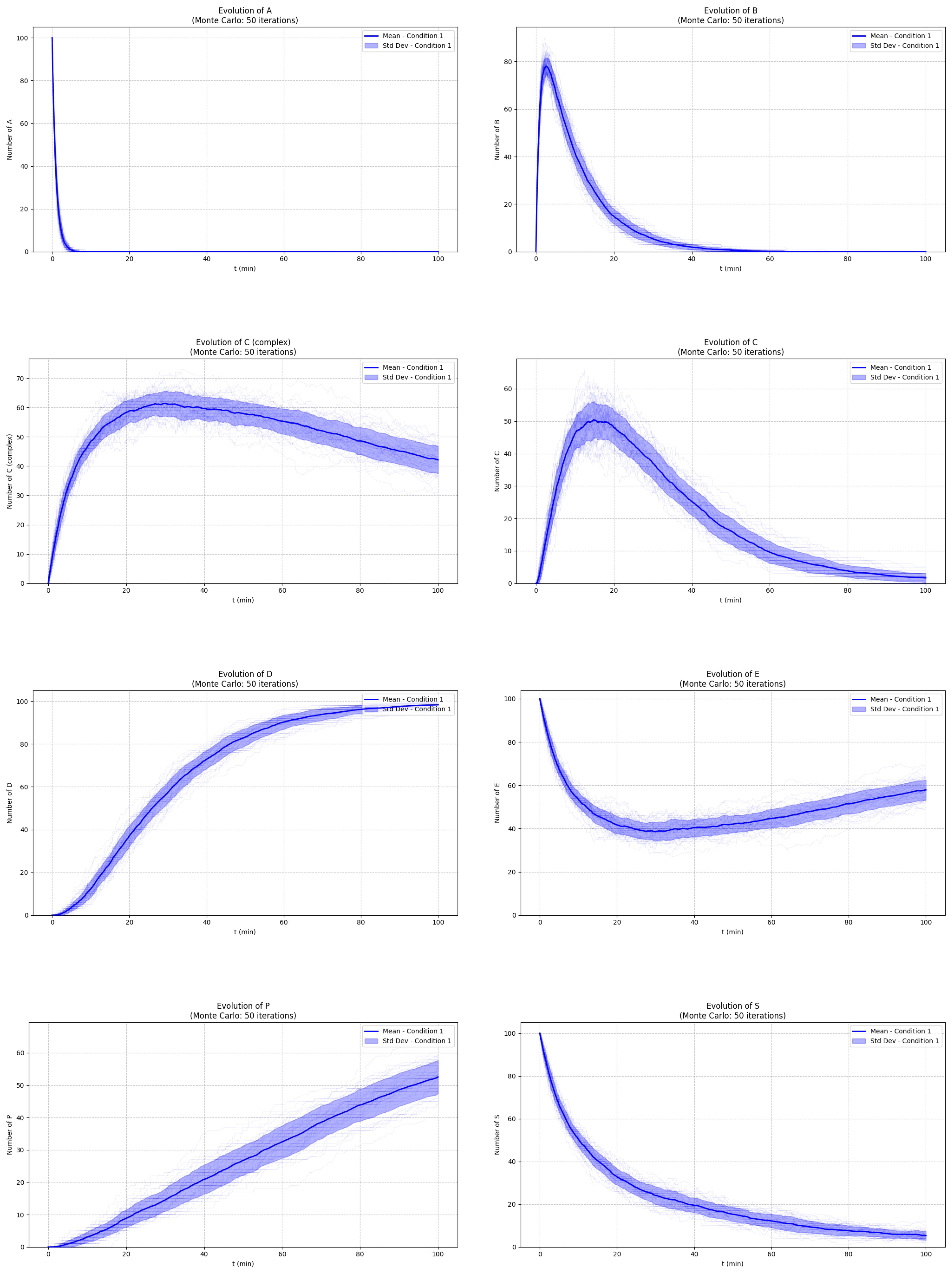}
    \caption{Results of each species}
    \label{fig:paper13-collage}
\end{figure}

\begin{figure}[H]
    \centering
    \begin{subfigure}{0.48\textwidth}
        \centering
        \includegraphics[width=\textwidth]{Appendix/Paper13/evolution_all_species.png}
        \caption{Result of all species}
        \label{fig:paper13-evolution-all-species}
    \end{subfigure}
    \hfill
    \begin{subfigure}{0.48\textwidth}
        \centering
        \includegraphics[width=\textwidth]{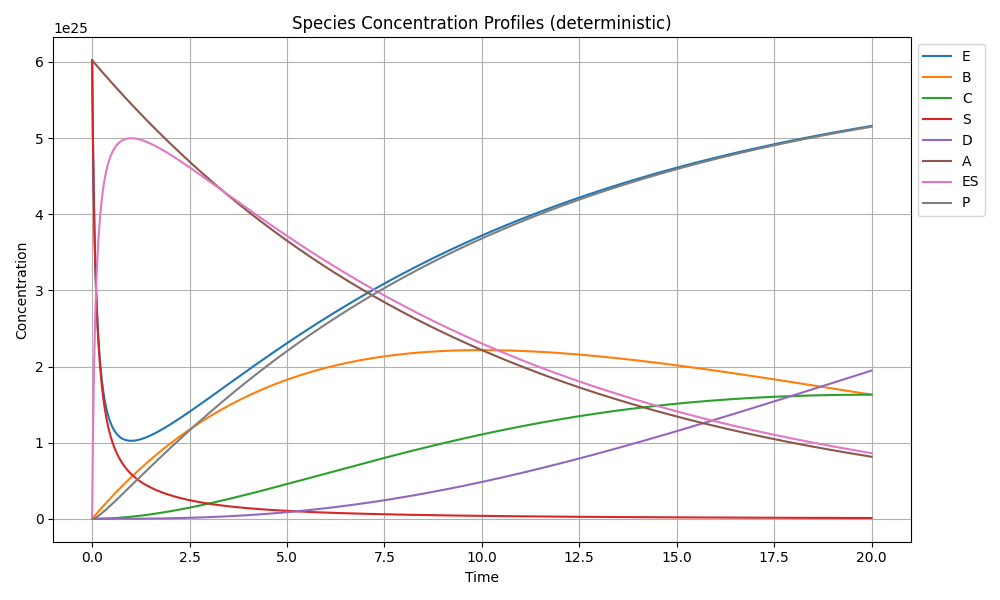}
        \caption{Result from Copasi (Deterministic)}
        \label{fig:paper13-copasi-deterministic}
    \end{subfigure}
    \vspace{0.5cm}
    \caption{Evolution and deterministic results from Copasi.}
    \label{fig:paper13-evolution-deterministic}
\end{figure}

\begin{figure}[H]
    \centering
    \begin{subfigure}{0.48\textwidth}
        \centering
        \includegraphics[width=\textwidth]{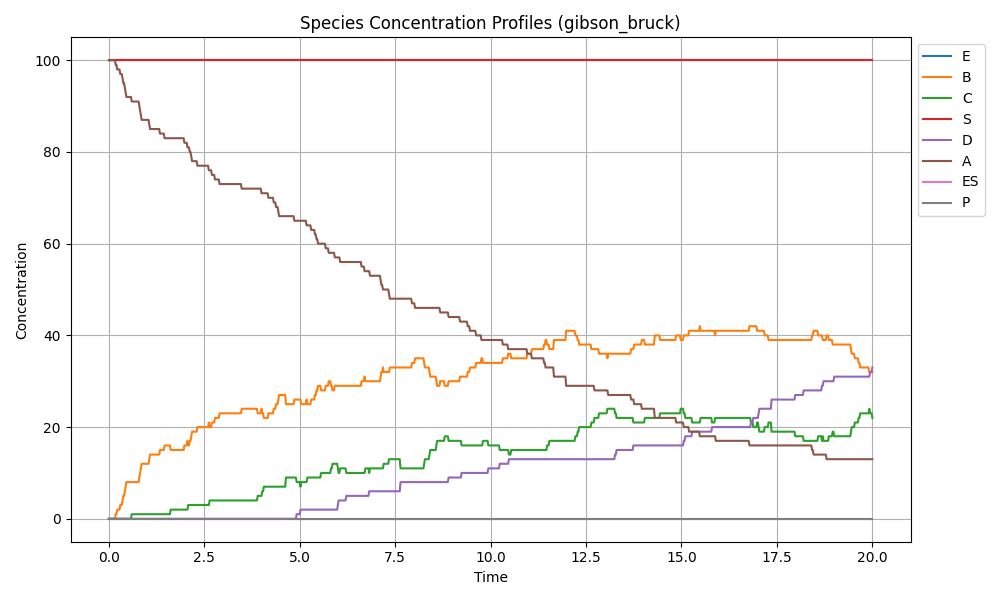}
        \caption{Result from Copasi (Stochastic Gibson Bruck)}
        \label{fig:paper13-copasi-stochastic-gibson}
    \end{subfigure}
    \hfill
    \begin{subfigure}{0.48\textwidth}
        \centering
        \includegraphics[width=\textwidth]{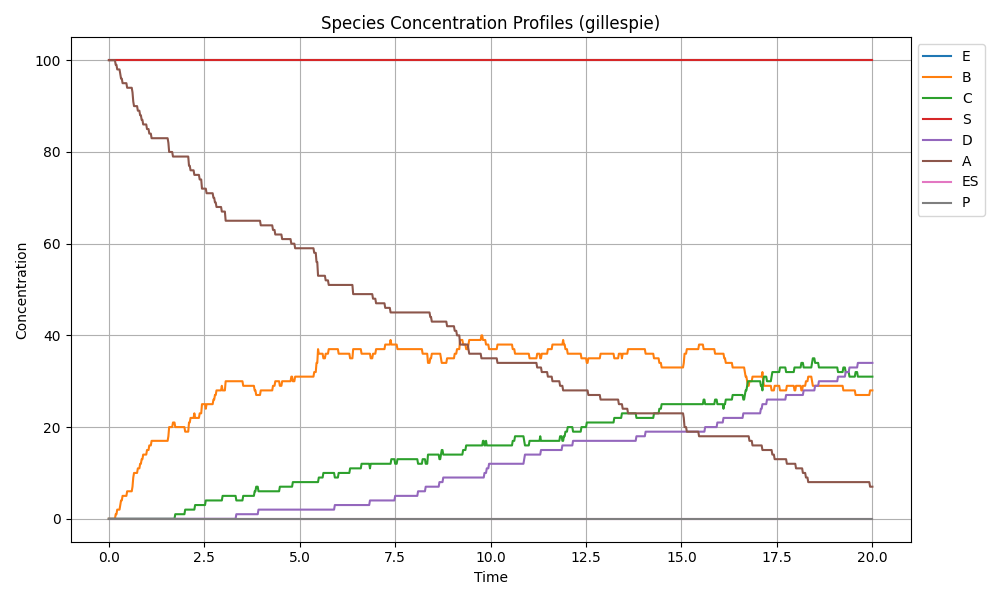}
        \caption{Result from Copasi (Stochastic Gillespie)}
        \label{fig:paper13-copasi-stochastic-gillespie}
    \end{subfigure}
    \vspace{0.5cm}
    \caption{Stochastic simulation results from Copasi.}
    \label{fig:paper13-stochastic-results}
\end{figure}

\end{document}